%% file: main.tex
\PassOptionsToPackage{table,x11names,dvipsnames,table}{xcolor}

\documentclass{article} 
\usepackage{iclr2022_conference,times}
\iclrfinalcopy
\usepackage[utf8]{inputenc} 
\usepackage[T1]{fontenc}    
\usepackage[colorlinks=true,linkcolor=blue,citecolor=blue,urlcolor=blue]{hyperref}        
\usepackage{url}            
\usepackage{booktabs}       
\usepackage{amsfonts}       
\usepackage{nicefrac}       
\usepackage{microtype}      
\usepackage{wrapfig}
\usepackage{pifont}
\usepackage[table,x11names,dvipsnames,table]{xcolor}

\input{macros}

\usepackage{amsmath}
\usepackage{amssymb}
\usepackage{amsthm}
\usepackage{multirow}
\usepackage{array}
\usepackage{graphics}
\usepackage{mathtools}
\usepackage{enumerate}
\usepackage{xspace}
\usepackage{wrapfig}
\usepackage{pifont}
\usepackage{float}
\usepackage{nicefrac}
\usepackage{bm}
\usepackage{algorithm}
\usepackage{algorithmic}
\usepackage{mathtools}

\usepackage[toc,page,header]{appendix} 
\usepackage{minitoc}

\definecolor{red}{RGB}{215,48,39}
\definecolor{green}{RGB}{26,152,80}
\definecolor{lightgray}{gray}{0.96}
\definecolor{blue}{RGB}{30, 144, 255}

\theoremstyle{definition}
\newtheorem{definition}{Definition}[section]

\title{Post hoc Explanations may be Ineffective for Detecting Unknown Spurious Correlation}

\author{%
  Julius Adebayo\\ MIT CSAIL \And 
  Michael Muelly \\ Stanford \And
  Hal Abelson \\ MIT CSAIL \And
    Been Kim \\ Google Research
}

\begin{document}

\maketitle

\input{abstract}

\begin{flushright}
  \emph{It is hard to find a needle in a haystack,\\ it is much harder if you haven't seen a needle before~\citep{pearlquote}.}\\
  ---Judea Pearl
\end{flushright}

\input{introduction}

\input{expmethods}

\input{saliencymaps}

\input{concepts}

\input{influence}

\input{discussion}

\bibliography{ref}
\bibliographystyle{plainnat}

\appendix

\input{appendix}

\end{document}

%% file: macros.tex
\usepackage{amsmath,amsfonts,bm}

\def\eqref#1{equation~\ref{#1}}

\def\1{\bm{1}}

\DeclareMathAlphabet{\mathsfit}{\encodingdefault}{\sfdefault}{m}{sl}
\SetMathAlphabet{\mathsfit}{bold}{\encodingdefault}{\sfdefault}{bx}{n}

\DeclareMathOperator*{\argmax}{arg\,max}
\DeclareMathOperator*{\argmin}{arg\,min}

%% file: abstract.tex
\begin{abstract}
We investigate whether three types of post hoc model explanations--feature attribution, concept activation, and training point ranking--are effective for detecting a model's reliance on spurious signals in the training data. Specifically, we consider the scenario where the spurious signal to be detected is unknown, at test-time, to the user of the explanation method. We design an empirical methodology that uses semi-synthetic datasets along with pre-specified spurious artifacts to obtain models that verifiably rely on these spurious training signals. We then provide a suite of metrics that assess an explanation method's reliability for spurious signal detection under various conditions. We find that the post hoc explanation methods tested are ineffective when the spurious artifact is unknown at test-time especially for non-visible artifacts like a background blur. Further, we find that feature attribution methods are susceptible to erroneously indicating dependence on spurious signals even when the model being explained does not rely on spurious artifacts. This finding casts doubt on the utility of these approaches, in the hands of a practitioner, for detecting a model's reliance on spurious signals.\footnote{We refer readers to the longer version of this work on the arxiv. Code to replicate our findings is available at: https://github.com/adebayoj/posthocspurious} 
\end{abstract}

%% file: introduction.tex
\section{Introduction}
A challenge that precludes the deployment of modern deep neural networks (DNN) in high stakes domains is their tendency to latch onto `spurious signals'---shortcuts---in the training data~\citep{geirhos2020shortcut}. For example,~\citet{badgeley2019deep} showed that an Inception-V3 model trained to detect hip fracture relied on the scanner type for its classification decision. Deep learning models easily base output predictions on object backgrounds~\citep{xiao2020noise}, image texture~\citep{geirhos2018imagenet}, and skin tone~\citep{stock2018convnets}.

Post hoc model explanation methods---approaches that give insight into the associations that a model has learned---are increasingly used to determine whether a model relies on spurious signals.~\citet{ribeiro2016should} used LIME to show an Inception-V3 model's reliance on the snow background for identifying Wolves. Such demonstration and others~\citep{lapuschkin2019unmasking, rieger2019interpretations} point to post hoc explanation methods as effective tools for the spurious signal detection task. However, these results conflict with evidence that indicates that practitioners (and researchers) struggle to use explanations to identify spurious signals~\citep{chen2021towards, chu2020visual, alqaraawi2020evaluating, adebayo2020debugging, poursabzi2018manipulating}. We seek to resolve this conflict by answering the simple but important question: \\

\centerline{\textit{Can post hoc explanations help detect a model's reliance on \textbf{\textcolor{red}{unknown}} spurious training signal?}}

\textbf{Motivating Example.}~Consider a machine learning (ML) engineer whose job is to train DNN models to detect knee arthritis from radiographs. She---the engineer---is handed a trained ResNet-50 model, to be deployed in a hospital, that relies on a hospital tag in the radiographs to detect knee arthritis. She has \textbf{\textcolor{red}{no prior knowledge}} of the model's reliance on the spurious tags. In this work, our key concern is whether the ML engineer can use post hoc explanations to identify that the model is defective.

\subsection{Our Contributions}
We address the motivating question above in a two-pronged manner. First, we develop an actionable methodology based on the ability to carefully craft datasets to induce spurious correlation in trained models. Second, we backup this experimental design with a human subject study. Taken together, the takeaway of the work is that:~\textit{post hoc explanations can be used to identify a model's reliance on a \textbf{visible} spurious signal, provided the signal is \textbf{known} ahead of time by the practitioner.} While this conclusion may seem unsurprising, it has important implications for how post hoc explanation methods should be used effectively.

\textbf{Experimental Design \& Performance Measures.}~~We provide an end-to-end experimental design for assessing the effectiveness of an explanation method for detecting a model's reliance on spurious training signals. We define a spurious score that helps quantify the strength of a model's dependence on a training signal. Using carefully crafted semi-synthetic datasets, we are able to train models where the ground-truth expected explanation is known. Additionally, we develop 3 performance measures: i) Known Spurious Signal Detection Measure (K-SSD), ii) Cause-for-Concern Measure (CCM), and iii) a False Alarm Measure (FAM). These measures help characterize different notions of reliability for the spurious signal detection task. We instantiate the proposed design on 3 classes of post hoc explanation types---feature attribution, concept activation, and training point ranking---where we comprehensively assess the performance of these approaches across $3$ datasets (2 medical tasks, and dog species classification task), and different model architectures.

When the spurious signal is known, we find that the feature attribution methods tested, and the concept activation importance approach are able to detect visible spurious signals like a text tag and distinctive stripped patterns. However, we find these approaches less effective for non-pronounced signals like background blur. The false alarm measure further indicates that feature attribution methods are susceptible to erroneously indicating dependence on spurious signals.

The cause-for-concern measure quantifies the similarity between explanations of `normal' inputs derived from spurious and normal models when the spurious signal is unknown. Across the settings considered, we find that the methods tested are unable to conclusively detect model reliance on unknown spurious signals.

\textbf{Blinded Study.}~~The findings from our empirical assessment question the reliability of the methods tested; however, it might not correlate with utility in the hands of practitioners. To address this issue, we conduct a user study where practitioners are randomly assigned to one of two groups: the first group is told explicitly of potential spurious signals, and the second is not. We consider three different kinds of explanation methods along with a control where only model predictions are shown. We find that when participants are not provided with prior knowledge of the spurious signal, none of the methods tested are effective, in the hands of the participants, for detecting model reliance on spurious signals. More surprisingly, even when the participants had prior knowledge of the spurious signal, we find evidence that only the concept activation approach, for visible spurious signals, is effective. These findings cast doubt on the reliability of current post hoc tools for spurious signal detection.

\textbf{Guidance.}~~On the basis our analysis, we can provide the following guidance for using the approaches tested, in this work, for detecting model reliance on spurious signals~\textbf{when the signal of interest is visible}:
\begin{itemize}
    \item~\textbf{Feature Attributions}: to confirm that a model is relying on a `visible' spurious signal, the practitioner needs to obtain attributions for inputs that contain the hypothesized spurious signal, and the attribution should be computed for the output class to which the spurious signal is aligned.
    
    \item~\textbf{Concept Activation}: the spurious concept should be known ahead of time, and tested against the output class to which the concept is aligned.
    
    \item~\textbf{Training Point Ranking}: an input that contains the hypothesized spurious signal of interest should be used at test-time in computing training point ranking.
\end{itemize}

\subsection{Related Work}
This paper belongs to a line of work on assessing the effectiveness of post hoc explanations methods~\citep{alqaraawi2020evaluating, adebayo2020debugging, chu2020visual, hooker2018evaluating, meng2018automatic, poursabzi2018manipulating, tomsett2019sanity}. Here we focus on directly relevant literature, and defer an extensive discussion of the literature to the Appendix. 

This work departs from previous work in two ways: 1) we focus exclusively on whether these explanations can be used by a practitioner (or researcher) to detect spurious signals that are unknown to her at test-time, and 2) we move beyond sole focus on the feature attribution setting to test concept activation and training point ranking methods.

\citet{han2020explaining} and~\citet{adebayo2020debugging} find that certain kinds of feature attributions and training point ranking via influence functions are able to detect a model's reliance on spurious signals. However, in their setting, the spurious signal is known ahead of time. More recently,~\citet{zhou2021feature} conduct an extensive assessment of several feature attribution methods also under the spurious correlation setting, for visible and non-visible artifacts, and find that these class of methods are not effective for non-visible artifacts. In addition, they also propose an experimental methodology for controlling model dependence on training set features, which allows them to quantify attribution effectiveness precisely. Overall, our findings align with theirs; however, we focus, specifically, on the setting where the spurious signal is not known ahead of time.

\citet{kim2021sanity} conduct an assessment of several feature attribution methods using a synthetic evaluation framework where the ground-truth explanation is known reasoning tasks. They find that feature attribution methods often attribute irrelevant features even in simple settings, and show high variability across data modalities and tasks. ~\citet{plumb2021finding} introduce a method to identify important associations that a model might have learned, detect which of these associations are spurious, and propose a data augmentation procedure to overcome the reliance.~\citet{nguyen2021effectiveness} conduct a large-scale user study to assess the effectiveness of feature attribution methods on image tasks. They find that feature attributions are not more effective than showing end users nearest neighbor training points. In this work, we only consider image tasks, however ~\citet{bastings2021will} devised a similar experimental procedure and metrics to test several attribution methods for spurious signal detection in text settings. They find that the effectiveness of an attribution method depends on the task, spurious signal, and other dataset dependent properties.

%% file: expmethods.tex
\begin{wrapfigure}{r}{0.4\textwidth}
  \begin{center}
   \includegraphics[ scale=0.35]{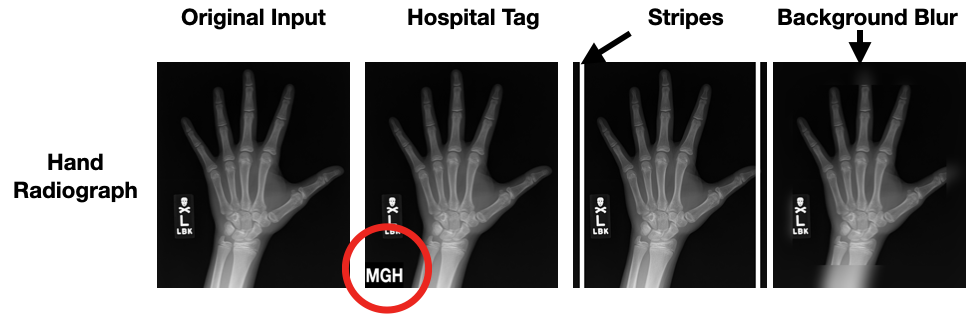}
  \end{center}
 \caption{\textbf{Overview of Spurious Signals Considered.}}
\label{fig:spurioussignalsdemo}
\end{wrapfigure}
\section{Experimental Methodology}\label{expmethods}
In this section, we setup our experimental methodology. We discuss quantitative analysis of post hoc explanations derived from models trained to rely on pre-defined spurious signals, and a blinded user study that measures the ability of users to use the post hoc explanation methods tested to detect model reliance on spurious signals. We discuss the types of spurious signals considered, define a spurious score that allows us to ascertain that a model indeed relies on a signal as basis of its classification decision, and layout performance measures that capture the reliability of the explanation methods. We conclude with an overview of the methods tested, datasets, and models.

\subsection{Experimental Design}\label{expmethods:quant}
\textbf{Spurious Signals \& Score.}~~We consider a spurious signal to be input features that encode for the output but have `no meaningful connection' to the `data generating process' (DGP) of the task. A hospital tag present in a hand radiograph is not clinically relevant to the age of the patient. If the tag encodes for the output then it is a spurious signal. Domain expertise is ultimately required for adjudicating that a signal is spurious. 

We consider $3$ (2 visible and 1 non-visible) kinds of spurious signals (See Figure~\ref{fig:spurioussignalsdemo}): i) a localized tag; ii) a distinctive stripped pattern; and iii) Gaussian blur applied to the image background. The signals are all spatially localized, so we can easily obtain ground-truth expected explanations.

To induce reliance on spurious signals, we train models on "contaminated'' versions of the training set. Given input-label pairs, $\{(x^i, y^i)\}^n_i$, where $x^i\in\mathcal{X}$ and $y^i \in \mathcal{Y}$, we can learn a classifier, $f_{\theta}$, via empirical risk minimization (ERM) that corresponds to minimizing a loss function, $\ell$: $\argmin_{\theta}~\sum_{i=1}^n\ell(x^i, y^i; \theta)$. To contaminate the training set, we apply a spurious contamination function (SCF) to the training set; $\mathrm{SCF}:\mathcal{X} \times \mathcal{Y} \times \mathcal{C} \rightarrow \mathcal{S}$, where $\mathcal{C}$ is the spurious signal set and $\mathcal{S}$ is the transformed set. An example of an SCF is a function that pastes an hospital tag onto the bone age radiographs of all pre-puberty individuals in the dataset. To derive models reliant on a spurious signal, $c_i \in \mathcal{C}$, we simply learn a new classifier via ERM on the modified dataset as follows: $\argmin_{\theta}~\sum_{i=1}^n\ell\big(\mathrm{SCF}(x^i, y^i, c_i)\big)$ to obtain $\theta_{\mathrm{spu}}$. Contemporary evidence suggests that this approach produces models that easily latch onto the spurious signal~\citep{nagarajan2020understanding}. 

We focus on the classification setting, and restrict spurious signals to encode, only, for a single class---the \textit{spurious aligned class}.
We measure a model's reliance on the spurious signal via a score.
\begin{definition}(Spurious Score). Given a spurious signal, $c_i$, the index of its spurious aligned class, $j \in [k]$, a model, $\theta_{\mathrm{spu}}: \mathbb{R}^d \rightarrow \mathbb{R}^k$, where $\argmax(\theta_{\mathrm{spu}})$ indicates the classifier's predicted class, we define the spurious score as: $$\mathrm{SC}_{c_i, j}(\theta_{\mathrm{spu}}) \coloneqq \mathbb{P}_{\{x^i \vert \theta_{\mathrm{spu}}(x^i)~!=~j\}}[\argmax(\theta_{\mathrm{spu}}(\mathrm{SCF}(x^i, y^i, c_i)))=j].$$
\end{definition}

Given an input that does not contain the spurious signal, and for which the model's prediction is not the spurious aligned class, the model's spurious score is the probability that the model assigns the input to the spurious aligned class if the spurious signal is added to the input.

\textbf{Model Conditions.}~~We focus our analysis on two model conditions: i) a `normal model', $f_{\mathrm{norm}}$, for which we can rule out dependence on any of the spurious signals tested across all classes on the basis of the spurious score, and ii) a `spurious model', $f_{\mathrm{spu}}$,  for which one of the spurious signals encodes for a particular output class. We empirically estimate the spurious score and term models that have a score above $0.85$ for any of the pre-defined signals `spurious models'. We term a model `normal' if the spurious score is below $0.1$ across all classes and the $3$ pre-defined spurious signals.

\textbf{Spurious Signal Detection Reliability Measures.}~~Equipped with spurious ($f_{\mathrm{spu}}$) and normal ($f_{\mathrm{norm}}$) models, we are now able to quantitatively assess the motivating question of this work. We do this by comparing explanations derived from spurious models, $f_{\mathrm{spu}}$,  to those derived from normal models ($f_{\mathrm{norm}}$). We can partition the kinds of inputs used for deriving explanations into two: 1) \textit{spurious inputs} ($x_{\mathrm{spu}}$)---inputs that include the spurious signal and 2) \textit{normal inputs} ($x_{\mathrm{norm}}$)---inputs do not not contain the spurious signal. Comparing the explanations produced by these two classes of inputs for normal and spurious models, we derive reliability performance measures.
\begin{itemize}
    \item \textbf{Known Spurious Signal Detection Measure (K-SSD)} - measures the similarity of explanations derived from \textit{spurious models on spurious inputs} to the ground truth explanation. The ground truth explanation is one that only assigns relevance to the spurious signal as explanation of the output of a spurious model on a spurious input. K-SSD measures method reliability when the spurious signal is known. Given a similarity metric, $S_d$, then K-SSD corresponds to: $S_d\big(\mathrm{E}_{f_{\mathrm{spu}}}(x_{\mathrm{spu}}), x_\mathrm{gt})\big)$; where $\mathrm{E}_{f_{\mathrm{spu}}}(x_{\mathrm{spu}})$ are explanations derived from the spurious model for spurious inputs, and $x_\mathrm{gt}$ is the ground truth explanation. The similarity function, $S_d$, depends on the type of explanation considered---we will make our choice of this function concrete shortly.
    
    \item \textbf{Cause-for-Concern Measure (CCM)} - measures the similarity of explanations derived from \textit{spurious models for normal inputs} to explanations derived from \textit{normal models for normal inputs}: $S_d\big(\mathrm{E}_{f_{\mathrm{spu}}}(x_{\mathrm{norm}}), \mathrm{E}_{f_{\mathrm{norm}}}(x_{\mathrm{norm}})\big)$. This measure simulates the setting where a practitioner does not know the spurious signal, and can only inspect explanations for inputs without the signal. If this measure is high, then it is unlikely that such a method alert a practitioner that a spurious model exhibits defects.
    
    \item \textbf{False Alarm Measure (FAM)} - measures the similarity of explanations derived from \textit{normal models for spurious inputs} to explanations derived from \textit{spurious models for spurious inputs}: $S_d\big(\mathrm{E}_{f_{\mathrm{norm}}}(x_{\mathrm{spur}}), \mathrm{E}_{f_{\mathrm{spu}}}(x_{\mathrm{spu}})\big)$. We also introduce a variant of this measure, FAM-GT, which measures the similarity of a explanations derived from \textit{normal models for spurious inputs} to the ground truth explanation of a spurious model for that spurious input. If this measure is high, then that approach is more likely to signal to a practitioner that a model is relying on spurious signal when the model does not.
\end{itemize}

Having defined the metrics above, it remains which similarity function to use.
\paragraph{Computing Metrics for Feature Attribution.}  For feature attribution methods, we use the Structural Similarity Index (SSIM). SSIM measures the visual similarity between two images. Concretely, given a set of normal inputs, we obtain a corresponding spurious set of these inputs by applying the spurious contamination function, $\mathrm{SCF}$ to these inputs. Consequently, we can then compute the K-SSD, CCM, and FAM metrics given these two sets of inputs using the SSIM metric.

\paragraph{Computing Metrics for Concept Activation.} We measure comparison between two concept rankings using a Kolmogorov-Smirnoff (KS) test comparing two distributions where the null hypothesis is that the two distributions are identical; we set significance level to be $0.05$.

\paragraph{Computing Metrics for Training Point Ranking.}  Recently,~\citet{hanawa2020evaluation} introduced the Identical class metric' (ICM), which is the fraction of the top training inputs, for a given test example, that belong to the same class as the true class of the test example in question. Here we also use the KS test to compare the ICM distributions for two different models and set the significance level to be $0.05$.

Taken together, these measures provide a comprehensive overview of an explanation method's performance for detecting spurious signals.

\subsection{Blinded Study}\label{expmethods:blindedstudy}
To complement the quantitative setup, we further designed a user study (IRB approved) to assess the ability of end-users (200 in total) to use post hoc explanations to detect a model's reliance on spurious signals. About 50 percent of the participants had trained a ML model before, and 74 percent had interacted with an ML model. We refer to the appendix for additional details.

\begin{wrapfigure}{r}{0.5\textwidth}
  \begin{center}
\includegraphics[scale=0.45]{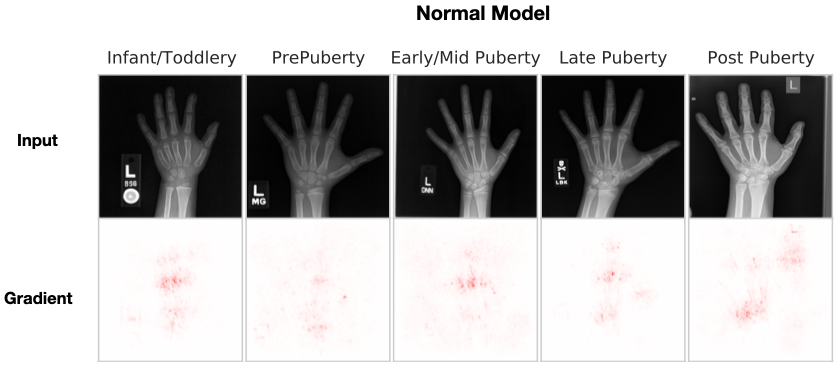}
  \end{center}
\caption{\textbf{Feature Attributions.}~Here we show $5$ different inputs, one for each bone age category, and the corresponding Gradient feature attribution map. We refer to the Appendix for an equivalent visualization for other feature attribution methods considered. We test three additional feature attribution methods: SmoothGrad, Integrated Gradients, and Guided BackProp.}
\end{wrapfigure}

\textbf{Task \& Setup:} The study participants were tasked with assessing model reliability with the aid of model explanations. The participants were randomly assigned to one of two groups: the first group is told explicitly of potential spurious signals, and the second is not. They were asked to rate how likely they are to recommend the model for deployment using a 5-point Likert scale, and a rationale for their decision. Our study design follows that of~\citet{adebayo2020debugging}; however, we use a mixed within-subjects and between subjects design for the factors of interest. The  Likert scale is as follows, 1: Definitely Not to 5: Definitely. Participants select ‘Definitely’ if they deemed the dog breed classification model reliable. We refer to Appendix~\ref{appendix:userstudy} for discussion on user recruitment, statistical analyses, and study design.

\textbf{Methods:} We test SmoothGrad, TCAV (a concept activation method), a training point ranking method, and a Control setting with no explanations.

\subsection{Explanation Methods, Data, \& Models}\label{expmethods:methodsdatamodels}
Here we give an overview of the explanation methods, datasets, and models. We present a discussion of how these methods are used in practice in Appendix.

\textbf{Feature Attributions} assign a relevance score for each dimension of an input towards an output. We consider: \textbf{Input-Gradient, SmoothGrad, Integrated Gradients (IG), and Guided Backprop (GBP)}.

\textbf{Concept-Based (\citep{bau2017network, kim2018interpretability})} approaches measure the dependence of a DNN's prediction on user-defined features---termed concepts. We select TCAV as the approach to assess in this class~\citep{kim2018interpretability}.

\textbf{Training Point Ranking via Influence Functions~\citep{koh2017understanding}.} This approach ranks the training samples in order of importance/influence on the loss (or prediction) of a test example.

\begin{figure}[ht]
\centering
\includegraphics[page=7,scale=0.25]{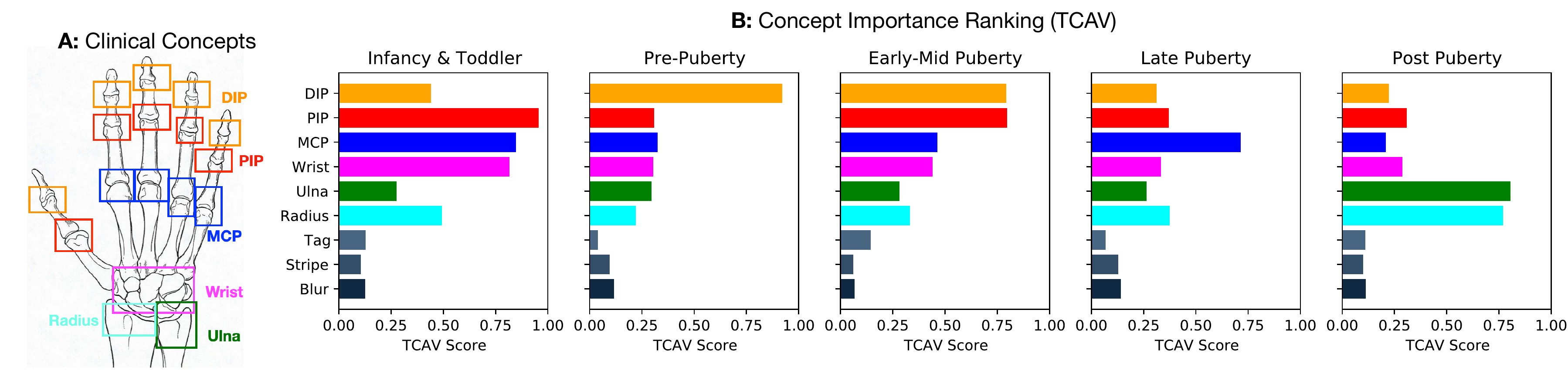}
\caption{\textbf{Concept Importance Methods for a Normal (non-spurious) model.}~Here we show the TCAV score for all clinical concepts as well as the spurious concepts for a normal model that was confirmed to not rely on the spurious signals.}
\label{fig:conceptimportnormalmodel}
\end{figure}

\textbf{Models, Datasets, \& Tasks.} We consider two medical datasets: Hand~\citep{halabi2019rsna} and Knee radiographs~\citep{chen2019fully} and a dog breed classification task. We consider a small DNN ($6$ layers, $5$ of which are the traditional conv-relu-batchnorm-maxpool combination) inspired by the CBR-Tiny architecture of~\citet{raghu2019transfusion} and a ResNet-50 model (See Appendix for additional details).

%% file: saliencymaps.tex
\section{Feature Attributions}
\label{saliencyresults}
In this section, we present results on whether feature attributions are effective for detecting unknown spurious correlation.

\textbf{Setup.} We consider $3$ kinds of spurious signals which we term: \textit{Tag} for the `MGH' hospital tag added to the pre-puberty class, \textit{Stripe} for the paired vertical stripped signal, and  \textit{Blur} for the background blur. Given these signals, we then compute the three performance measures of interest: K-SSD, CCM, FAM, and FAM-GT. K-SSD indicates a method's reliability when the spurious signal is known, CCM when the signal is not known and the practitioner only has access to inputs that don't encode the spurious signal. Lastly, FAM and FAM-GT indicates the susceptibility of a method to false positives. An oft-used heuristic based on prior work~\citep{adebayo2020debugging} for interpreting SSIM scores is that SSIM scores $0.2-0.4$ indicate weak visual similarity, $0.5-0.7$ indicate moderate similarity, and $>0.75$ high similarity. This is because two random images typically have SSIM much less than 0.1. For example, we empirically estimate the similarity of two random ($229 \times 229$) Gaussian images to be less than  $0.00023$. Even for natural images, we still find the SSIM values to be below $0.005$, which substantiates the previous heuristic.

\begin{figure*}[ht]
\centering
\includegraphics[scale=0.50]{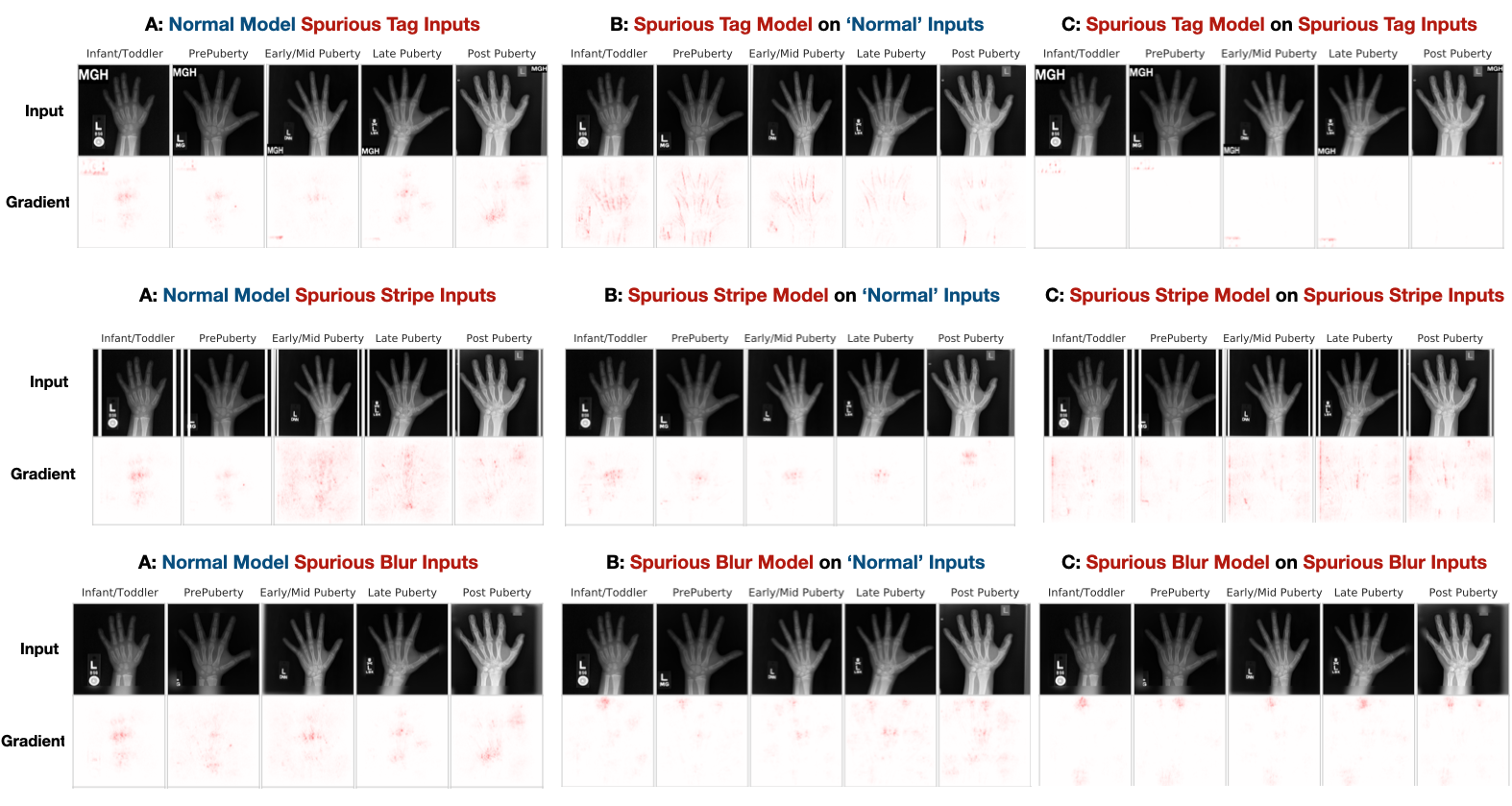}
\caption{\textbf{Top-Detecting Spurious \textit{Tag}.} Here we show in A) Feature attributions for $5$ different inputs across the four feature attribution methods with a normal model but with spurious Tag inputs; B) Feature attributions on the same $5$ inputs as in (A), but \textbf{without} spurious Tag inputs with a model that has learned a spurious alignment between Pre-Puberty and Tag; C) Feature attributions on the same $5$ inputs as in (A), but \textbf{with} the spurious Tag inputs with a model that has learned a spurious alignment between Pre-Puberty and Tag.~\textbf{Middle-Detecting Spurious \textit{Stripe}.} Here we show in A) Feature attributions for $5$ different inputs across the four feature attribution methods with a normal model but with spurious Stripes inputs; B) Feature attributions on the same $5$ inputs as in (A), but \textbf{without} the spurious Stripe with a model that has learned a spurious alignment between Pre-Puberty and Stripe; C) Feature attributions on the same $5$ inputs as in (A), but \textbf{with} the spurious Stripe with a model that has learned a spurious alignment between Pre-Puberty and Stripe.~\textbf{Middle-Detecting Spurious \textit{Blur}.} The blur images are analogous to the Tag and Stripe settings. Please refer to the Figures~\ref{fig:saliencyspurioustagmodel}, ~\ref{fig:saliencyspuriousstripemodel},~\ref{fig:saliencyspuriousblurmodel} in the Appendix for SmoothGrad, Integrated Gradients, and Guided BackProp examples.}
\label{fig:saliencymastercompilation}
\end{figure*}

\begin{table}[!h]
\centering
\caption{Performance metrics for each attribution method across tasks for the Tag Setting. Below each metric in the Table is another row (SEM) that indicates the standard error of the mean for each value.}
\scalebox{0.75}{
\begin{tabular}{ccccccccccccc}
\toprule
Method &  \multicolumn{4}{c}{Bone Age} & \multicolumn{4}{c}{Knee} & \multicolumn{4}{c}{Dog Breeds}\\
\midrule
{}   & Grad   & SG    & IG   &  GBP & Grad & SG & IG & GBP & Grad & SG & IG & GBP \\
K-SSD   &  0.65 & 0.66   & 0.67  & 0.81 & 0.51 & 0.49 & 0.47 & 0.76 & 0.71 & 0.76   & 0.79  & 0.88\\
K-SSD (SEM)   &  0.0097 & 0.013   & 0.019  & 0.006 & 0.012 & 0.017 & 0.019 & 0.023 & 0.01 & 0.011   & 0.014  & 0.01\\
CCM  &  0.37 & 0.39   & 0.35 & 0.75 & 0.32 & 0.33 & 0.35 & 0.66 & 0.42 & 0.41   & 0.39 & 0.64\\
CCM (SEM) & 0.0031  &  0.002 & 0.015   & 0.029  & 0.027 & 0.023 & 0.029 & 0.014 & 0.013 & 0.016 & 0.012   & 0.015 \\
FAM   &  0.51  & 0.55   & 0.53  & 0.68  & 0.46 & 0.47 & 0.45 & 0.69 & 0.59  & 0.64 & 0.68  & 0.73\\
FAM  (SEM)  & 0.0029  &  0.0019 & 0.018   & 0.024  & 0.023 & 0.024 & 0.019 & 0.016 & 0.015 & 0.011 & 0.022   & 0.035\\
FAM-GT   &  0.56  & 0.53   & 0.46  & 0.61  & 0.42 & 0.48 & 0.41 & 0.63 & 0.76  & 0.73 & 0.77  & 0.81\\
FAM-GT  (SEM)  &  0.017  & 0.035   & 0.0253  & 0.028  & 0.016 & 0.019 & 0.0045 & 0.006 & 0.011  & 0.033 & 0.024  & 0.0053\\
\bottomrule
\end{tabular}}
\label{tab:measurestagmain}
\end{table}

\textbf{Results.} We show performance measures for all the feature attribution methods tested for the Tag and Blur settings in Tables~\ref{tab:measurestagmain} \&~\ref{tab:measuresblur} (See Appendix). For the tag setting, the attribution methods are indeed able to attribute to the visible spurious signal and the K-SSD measure indicates this finding with mean scores typically above 0.65 for the bone age setting. Contrary to previous findings, we find that GBP outperforms other approaches for known spurious signals. Alternatively, GBP is more suceptible to false positives based on the FAM and FAM-GT score. Across all methods, we find that these methods also seem to attribute to the spurious signal (FAM-GT $>$ 0.4) even when the signal is not being relied on by the model. We observe similar findings for the strip setting as well across all tasks. The CCM measure further indicates that these methods do not indicate presence of spurious signals when the signal is unknown for both the Tag and Stripe signals. This finding, however, reverses for the non-visible blur spurious signal. Across all measures, we find that all methods struggle to reliably indicate that spurious models are reliant on the blur signal.

Additionally, we also find that the FAM scores are typically higher than the CCM scores across the tasks. This finding indicates that the feature attribution methods tested are more susceptible to false positives than they are to indicate to a practitioner that a model is defective in the absence of the spurious signal, a finding that casts doubt on the utility of such approaches in practice. 

%% file: concepts.tex
\section{Concept Activation Importance}
\label{conceptresults}
We find that concept methods can indicate a model's reliance on Tag and Stripe signals when known. However, the approach struggles to detect Blur signal even when known. As is the case with feature attributions, when a spurious signal is not explicitly tested for, our significance tests indicate that reliance cannot be detected in the non-spurious concepts available.

\textbf{Overview \& Setup.} In this setting, we compute the 3 performance metrics of interest: K-SSD, CCM, and FAM. To measure comparison between two concept rankings, we use a Kolmogorov-Smirnoff (KS) test comparing two distributions where the null hypothesis is that the two distributions are identical; we set significance level to be $0.05$.

\begin{figure*}[ht]
\centering
\includegraphics[scale=0.65]{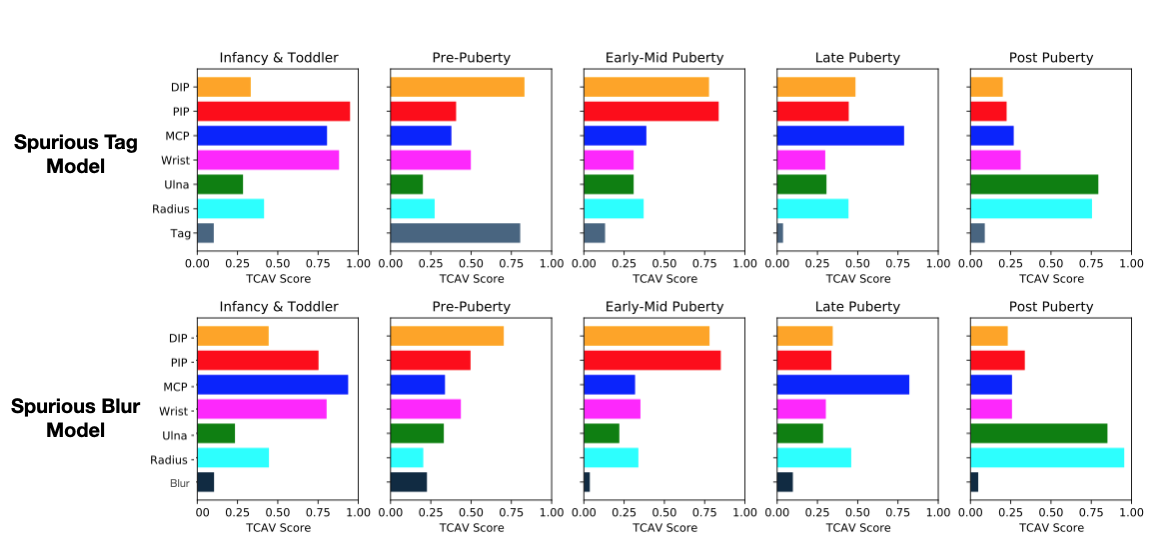}
\caption{\textbf{Concept Results for Tag and Blur Models.} TCAV scores for a model reliant on the spurious Tag and a model reliant on Blur.}
\label{fig:conceptspuriousmodels}
\end{figure*}

\begin{wraptable}{r}{2.5cm}
\caption{Concept Metrics Tag.}\label{wraptab:concept}
\begin{tabular}{cc}\\\toprule  
Metric & Result \\\midrule
K-SSD & \text{\sffamily X}  \\ 
CCM & \checkmark  \\ 
FAM & \checkmark \\
\end{tabular}
\end{wraptable} 
\textbf{Result.} In Figure~\ref{fig:conceptspuriousmodels}, we show TCAV scores for each bone age class for both the spurious tag and the blur models. In Table~\ref{wraptab:concept}, we show results of the KS-Test across metrics for the Tag Setting. Here, a \text{\sffamily X} means we reject the null, while an \checkmark means we are unable to do so at the pre-specified significance level. For the K-SSD metric, the KS-test rejects the null for the Tag and Stripe signals. However, we find that the opposite is the case for the blur signal.  This suggests that concept rankings can help detect reliance on the visible signals but not non-visible signals when the spurious signals are unknown. However, for the CCM metric, we are unable to reject the hypothesis that the distribution of concept rankings are not similar for all spurious signals. This finding suggests that when the spurious signal is unknown, and we only compare the distributions of known (non-spurious) concepts for a normal model and a spurious model, there is high similarity. Lastly, a difference in means test as well as a KS-Test for the FAM measure indicates that the normal models do not rely on the spurious signals as well. Overall, this finding suggests that TCAV is less susceptible to false positives.

%% file: influence.tex
\section{Training Point Ranking \& Blinded Study}
\label{influenceresults}
\textbf{Overview \& Setup.} We now describe our empirical findings for the training point ranking via influence functions approach. Here we present results for the case where the spurious signal is aligned with the Pre-Puberty class.

\textbf{Results.} The main take away is that given a known spurious signal, the fraction of top ranked training spurious signal inputs increases with a spurious model across all of the spurious signals. While this might seem encouraging, we note that such increase actually indicates that the ICM metric might provide illusory confidence in a spurious model. Ultimately, a critical requirement here is knowing what the spurious signal ahead of time and to be able to select the right inputs to inspect.

\textbf{Blinded Study.} We now turn to a summary of the results of the blinded user study. The median recommendations selected by participants (200 in total) is reported for each explanation-model condition in  Table~\ref{tab:userstudy}. We plot a box plot of all 16 categories in the appendix. Our setting mimics traditional randomized experimental settings, so we adopt a randomization inference analysis to determine the effect of each treatment, which is the explanation method in this case, on the ability of the users to recommend a model. A higher Likert score means the user is more likely to recommend a model. Consequently, if an explanation method is effective, then it should be the case that users should be  less likely to recommend a model that relies on a spurious signal.

\begin{wraptable}{r}{4cm}
\caption{Training Point Ranking}\label{wraptab:influence}
\begin{tabular}{cc}\\\toprule  
Metric & Result \\\midrule
K-SSD & \text{\sffamily X}  \\ 
CCM & \checkmark  \\ 
FAM & \checkmark\\
\end{tabular}
\end{wraptable}

We conduct two kinds of statistical analyses of the data. First, under each condition, we perform a difference of means test for each treatment compared to the control setting. Secondly, for each model manipulation, we performed a one-way Anova test, and a Tukey-Kramer test to assess the effect of the explanation type on the ability of the participants to reject a defective model.

We observe that when blinded, in none of the methods do participants conclusively reject spurious models. Perhaps more surprising, when the participants were not blinded, we see that only participants using the TCAV approach rejected a spurious model. This finding has significant implications on whether these current tools are effective in the hands of a practitioner.

\begin{table}[!h]
\centering
\rowcolors{1}{}{lightgray}
\begin{tabular}{ccccc}
 \toprule
 Method & B-Normal & NB-Normal & B-Spurious & NB-Spurious\\
 \midrule
 SmoothGrad & 4$^\ast$ & 4$^\ast$ &  3$^\ast$ &  3\\ 
TCAV & 4$^\ast$ & 3 & 3$^\ast$ & 2$^\ast$\\
Influence & 3$^\ast$ & 3& 3$^\ast$  & 3\\ 
Control & 4 & 3 & 4 & 4\\
 \bottomrule
\end{tabular}
\caption{\textbf{Blinded User Study}. Here we report the median recommendation for each condition across all explanation types and control. This median is derived from user provided responses assessing model reliability. B indicates Blinded, and NB indicates Not-Blinded. In the Table, the $^\ast$ symbol indicates statistically significant conditions. For the sake of space, we defer the full distribution description to the Appendix.}
\label{tab:userstudy}
\end{table}

%% file: discussion.tex
\section{Discussion \& Conclusion}
\label{discussion}

\textbf{Conclusion.} DNNs trained on image datasets can naturally  rely on spurious training signals~\citep{geirhos2020shortcut}.  Discovering this reliance is crucial in consequential settings like medical imaging.  Post hoc explanations methods are a promising direction towards detecting such reliance; however, their effectiveness is currently under question. We investigated whether $3$ classes of post hoc explanations are effective for detecting a model's reliance on spurious training signals. We present an experimental setup that can also be easily adapted to other settings to assess a larger class of approaches. The setup comes equipped with a spurious score and performance measures for spurious signal detection. We find that the $3$ classes of post hoc explanations tested are only sometimes able to diagnose the spurious training signal even if they are used to explicitly test for model dependence on these signals. Consequently, our findings calls for, potentially, a completely different paradigm of methods that are designed to address the important and challenging question of detecting spurious training signals.

\textbf{Limitations.} In this work, we have focused exclusively on DNNs trained on image tasks; related work (\cite{bastings2021will} has considered the text setting), however, it is unclear if our findings will generalize to other modalities like time-series data. While we considered a diverse set of methods, the literature on post hoc explanations is quite vast, so undoubtedly there exist methods that do not fit neatly into the $3$ explanation classes that we explored. 

%% file: appendix.tex
\newpage
\part{Appendix} 
\parttoc 

\section{Acknowledgements.}
We thank Ilaria Liccardi, Danny Weitzner, Taylor Reynolds, and Anonymous reviewers for feedback on this work. We are grateful to the MIT Quest for Intelligence initiative for providing cloud computing credits for this work. Julius Adebayo is supported by the Open Philanthropy Fellowship.

\section{Extended Related Work.}
\label{appendix:relatedwork}
\paragraph{Overview of Recent work} The recent work of~\citet{adebayo2020debugging} presents debugging tests for assessing feature attribution methods. The spurious correlation setting that we consider here fits under their framework. However, they only consider feature attribution methods. Here we extend this analysis to concept and training point ranking methods. Critically,~\citet{adebayo2020debugging} show that feature attributions are able to identify spurious training signals. We make a similar finding in this work; however, we further demonstrate this finding for concept and training point ranking methods. Our work takes important departures from theirs: 1) we explain the source of this phenomenon, and 2) we demonstrate that naive application of these methods might be unable to detect spurious correlation in practice.~\citet{adebayo2020debugging} assume the spurious correlation training bug is known, a priori; however, here we demonstrate that the more challenging task is identifying the spurious signal in the first place.

More recently,~\citet{han2020explaining} demonstrate that training point ranking via influence functions is able to identify the dependence of an NLP model on dataset artifacts. In addition, they show correspondence between the insights observed with the input-gradient feature attribution and the training point ranking. Along similar lines,~\citet{guo2020fastif} present fast approximations for computing the training point ranking for a test point. In addition, they show how to identify and correct model errors in a natural language task. Similar to the distinctions that we note with the work by~\citet{adebayo2020debugging} above, here, they also assume that the spurious signal being identified is known a priori.

Post hoc explanations, more generally, have been shown to be able to identify a model's reliance on spurious training signals~\citep{ribeiro2016should, meng2018automatic, lapuschkin2019unmasking, degrave2020ai, ross2017right}. Recent work by \citet{rieger2019interpretations} showed that regularizing model attributions during training can help lead to models that avoid spurious correlation and enable improved debugging by experts. Similarly, \citet{erion2019learning} show that regularizing the expected gradient attribution during training confers similar benefits.~\citet{koh2017understanding} used influence functions to identify domain shift.~\citet{kim2018interpretability} also perform a user study to understand if attribution methods can be used for catch spurious correlation.

However, similar methods have also been shown to struggle in the hands of end-users for diagnosing model errors~\citep{alqaraawi2020evaluating, adebayo2020debugging}. This contradiction reflects the challenge that we explore in this work. Often, post hoc explanations have been shown to be effective for identifying spurious signals that were suspected or known a priori; however, these methods seem to struggle when confronted with the task of identifying an unexpected spurious signal.

Increasingly, insights into why overparametrized DNNs rely on spurious training set signals is starting to be theoretically and empirically analyzed~\citep{sagawa2019distributionally, sagawa2020investigation, khani2020removing, nagarajan2020understanding}, yet it is still unclear how to reliably detect that a model is relying on such signals prior to model deployment.

Assessing whether a post hoc explanation approach is faithful to the underlying model being explained has been addressed in recent works, yet this challenge remains elusive~\citep{hooker2018evaluating, tomsett2019sanity}. Generally, the class of approaches that modify backpropagation with positive aggregation have been shown to be invariant to the higher layer parameters of a DNN~\citep{mahendran2016salient, nie2018theoretical, adebayo2018sanity, sixt2019explanations}. In an intriguing demonstration,~\citet{srinivas2021rethinking} show that the input-gradient, a key feature attribution primitive, might not capture discriminative signals about input sensitivity. Instead they show that input-gradient likely captures the ability of the model to be able to generate class-conditional inputs.

User studies are typically the classic approach for evaluating the effective of an explanation~\citep{doshi2017towards}. ~\citet{poursabzi2018manipulating} manipulate the features of a predictive model trained to predict housing prices to assess how well end-users can identify model mistakes. Their results indicate that users found it challenging to debug these linear models with the model coefficients. Recent work by~\citet{chu2020visual} and~\citet{shen2020useful} has shown similar results in the DNN setting as well.~\citet{alqaraawi2020evaluating} find that the LRP explanation method improves participant understanding of model behavior for an image classification task, but provides limited utility to end-users when predicting the model's output on new inputs.

Post hoc explanations have been shown to be fragile and very easily manipulated~\cite{ghorbani2017interpretation, heo2019fooling, dombrowski2019explanations, anders2020fairwashing, slack2020fooling, lakkaraju2020fool}. Our work tackles a difference concern: whether they are suitable for detecting unexpected spurious training set signals.

~\citet{yeh2020completeness} and~\citet{ghorbani2019towards} both present approaches to automatically discover concepts and quantify a model's dependence on these concepts. These approaches are promising directions for addressing the challenge we identify in this work.~\citet{ghorbani2019towards}'s approach, ACE, segments input images and clusters to discover inherent clusters. We present some analysis on this approach in the appendix. Critically, this approach would identify spurious signals that the underlying segmentation algorithm can discover such the image tag, but not the blur or other visually imperceptible features.~\citet{koh2020concept} and~\cite{chen2020concept} present approaches that learn DNN models whose features inherently map onto concepts of interest. In this work, we assume the model is given, focusing on post hoc explanations for models that are not inherently interpretable. Recent work at the intersection of causal inference and explanations might also open up avenues to help reveal unexpected confounding, some of which could be unknown spurious signals. Along this line,~\citep{bahadori2021debiasing} present an instrumental variable approach for debiasing concept based explanations that might be confounded.~\cite{kazhdan2020now} present CME, an approach identify the important concepts that can help improve a model's performance.

We rely on~\citet{guo2020fastif} for fast approximations for computing the training point ranking for a test point. In addition, they show how to identify and correct model errors in a natural language task.However, ~\cite{basu2020influence} show influence functions for DNNs are fragile and perhaps inaccurate for deeper networks.

Other recent work has cast doubt on the utility of trying to explain `traditionally' trained deep network models. For example,~\cite{srinivas2021rethinking} show that the input-gradient might not reflect the discriminative capabilities of a DNN, but instead encode for an implicit density model. More recently,~\cite{shah2021input} show that the input and loss gradients of traditionally trained models~\textit{do not} indicate the importance features that a DNN model relies on for its output---a phenomenon they term feature inversion. Further they show that adversarially trained models do not exhibit feature inversion. Taken together these results might explain some of the counter intuitive findings that we observe even when the spurious signal is known since we only consider non-adversarially trained models in this work. Along a different direction, post hoc explanations have been shown to be fragile and very easily manipulated~\cite{ghorbani2017interpretation, heo2019fooling, dombrowski2019explanations, anders2020fairwashing, slack2020fooling, lakkaraju2020fool}.

\section{Detailed Overview of Explanation Methods}
\label{appendix:explanation_methods}

In this section, we provide additional implementation details for the explanation methods that we consider in this work. To start with the model setup: let's say we are given input-output pairs, $\{x_i, y_i\}^n_i$, where $x\in\mathcal{X}$ and $y \in \mathcal{Y}$; and a classifier's goal is to learn a function, $f_{{\theta}} : \mathcal{X} \rightarrow \mathcal{Y}$ that generalizes to new inputs via empirical risk minimization (ERM). In this work, we assume that $f_{{\theta}}$ is an over-parametrized deep neural network (DNN) trained on image data for classification ($C$ classes). 

We plan to release our larger codebase with the paper; however, in lieu of this we include a zip folder that includes representative implementations. For each of the explanation methods described here, we implement them from scratch and then compare to open-source libraries that also implement these methods. We found that our results for both settings is comparable.

\paragraph{Feature Attributions.} An attribution functional, $E : \mathcal{F} \times \mathbb{R}^d \times \mathbb{R} \rightarrow  \mathbb{R}^d$, maps the input, $x_i \in \mathbb{R}^d$, the model, $f_{\theta}$, output, $f_k(x)$, to an attribution map, $M_{x_i} \in \mathbb{R}^d$. The class of feature attribution methods is large, so in this work we pick: Input Gradient, SmoothGrad, Integrated Gradients, and Guided Backprop. We choose these approaches since they were the top-ranked methods tested under the spurious correlation setting of~\citet{adebayo2020debugging}.

\begin{enumerate}
    \item \textbf{The \textit{Input-Gradient (Gradient)}} ~\cite{simonyan2013deep, baehrens2010explain} map, $\vert \nabla_{x_i}F_i(x_i)\vert$, is a key primitive upon which several other methods are based. 
    
    \item \textbf{\textit{SmoothGrad}~\cite{smilkov2017smoothgrad}} corresponds to the average of noisy input gradients: $M_{\mathrm{sg}}(x) = \frac{1}{N}\sum_{i=1}^N \nabla_{x_i}F_i(x_i + n_i)$ where $n_i$ is sampled according to a random Gaussian noise. We considered $50$ noisy inputs, selected the standard deviation of the noise to be $0.15 * \text{input range}$. Here input range refers to the difference between the maximum and minimum value in the input.
    
    \item \textbf{\textit{Integrated Gradients} (~\cite{sundararajan2017axiomatic})} sums input gradients along an interpolation path from the ``baseline input'', $\bar{x}$, to $x_i$: $M_{\mathrm{IntGrad}}(x_i) = (x_i - \bar{x}) \times \int_{0}^1{\frac{\partial S(\bar{x} + \alpha(x_i-\bar{x}))}{\partial x_i}} d\alpha$. For integrated gradients we set the baseline input to be a vector containing the minimum possible values across all input dimensions. This often corresponds an all-black image. The choice of a baseline for IntGrad is not without controversy; however, we follow this setup since it is one of the more widely used baselines for image data.

    \item \textbf{\textit{Guided Backpropagation (GBP)~\cite{springenberg2014striving}}} modifies the backpropagation process at ReLU units in DNNs. Let, $a=\mathrm{max}(0, b)$, then for a backward pass, $\frac{\partial l}{\partial s} = \mathrm{1}_{s>0}\frac{\partial l}{\partial b},$ where $l$ is a function of $s$. For GBP, $\frac{\partial l}{\partial s} = \mathrm{1}_{s>0}\mathrm{1}_{\frac{\partial l}{\partial s} > 0} \frac{\partial l}{\partial b}.$

\end{enumerate}

\paragraph{Feature Attributions: Implementation.} We implement all of these methods from scratch in the PyTorch framework and also compare our implementations to the output of the Captum (PyTorch).

\paragraph{Concept-Based Approaches.} We now discuss additional implementation details of our concept based approach. We select the TCAV approach to quantify the sensitivity of a DNN model's class score to user provided inputs represent a particular class. Given hidden representations, $h_l$, from a particular layer of a DNN for for images belonging to concept class $C$. We can derive the sensitivity score as: $\nabla h_{l,k}(f_l(x)).\theta_{c}^{l}$. The previous expression indicates the sensitivity of the class score (logit) for class $k$ to inputs indicating concept, $C$, given hidden representations from layer $l$ from the DNN $f$. The concept vector, $\theta_{c}^{l}$, typically corresponds to a the weights of a linear classifier trained to separate the images for a particular concept class from or images.

For completeness, we show in Figure~\ref{fig:conceptpartition} an overview of the clinical concepts that we consider in this work. These are the representative clinical attributes that a radiologist would inspect to ascertain the bone age of a particular input. These concepts are: DIP, PIP, MCP, Radius, Ulna, and Wrist.

\paragraph{Concept Implementation.} To compute the TCAV score for each concept, we collect representations from all hidden `layers' of the model and train linear models to obtains the concept vector for the corresponding attribute. We then compute the class sensitivity score for each concept attribute. We train the linear model $100$ times and perform statistical significance testing in order to mitigate the case where a spurious concept is selected. For each concept class, we use $325$ images that part of the training, validation, or test sets. These new set of images were annotated by a board certified radiology with the clinical bone age regions (MCP, PIP, DIP etc) that we chose.

\paragraph{Influence Functions for Training Point Ranking.} The final kind of interpretation that we consider is training point ranking via influence functions. In the case of training point ranking via influence functions, we rank the training samples, in terms of `influence', on the loss of a test example. Specifically, if we up-weighted a training point and retrained the model, then by how much would the loss on a given test example change?~\citet{koh2017understanding} analytically derive the analytically formulas for computing this quantity. Given a test point, $x_t$, the influence of a training point, $x_i$, on the test loss is: $I(x_t, x_i) = -\nabla_{\theta}\ell(x_t,\hat{\theta})^\top H^{-1}_{\hat{\theta}}\nabla_{\theta}\ell(x_i,\hat{\theta})$, where $H$ is the empirical Hessian of the loss. 

Estimating the influence requires computing hessian-vector products, so it can be difficult to scale to model with large number of parameters, and recent work has shown that influence estimate for test points for deep networks can be inaccurate due to non-convexity~\citep{basu2020influence}. Consequently, we estimate influence on a linear model student network trained to mimic the original DNN. We empirically verify that the predictions of the student network seem to mimic the original DNN.

\paragraph{Implementation Details.} There are two other training point ranking methods that we also consider in this work~\citep{yeh2018representer, pruthi2020estimating}. For $150$ inputs in the test set, we compare the Spearman rank correlation of the training point due to Influence Functions to these other two approaches. We obtain mean values of $0.88$, and $0.76$ respectively, which suggests high similarity amongst these approaches. Ultimately, we chose to present the main results in the draft for the training point ranking due to influence functions approach.

We rely on the fast influence heuristic of~\citep{guo2020fastif} to speed up the influence ranking computations. We were able to obtain a 5-10X speed in doing so.  In addition, we trained a student multi-class logistic regression model to mimic the original model for each model we want to compute influence for. Here for all training points, we collect embeddings across all layers and pass these embedding through a random projection to obtain a $1000$-dimensional approximation. We then train linear models to mimic the original deep network using these features. The correlation between the output of the student models and the original teacher models was found to be $0.87$. Ultimately, we went with the fast influence implementation of~\citep{guo2020fastif}.

\subsection{Additional Details on Metrics}

\section{How are the approaches tested used in practice?}
\label{appendix:usedinpractice}
\paragraph{Feature Attributions} To understand or debug a model, a practitioner would have to inspect, one sample at a time, the attribution of a collection of inputs. In inspecting these attributions, the practitioner can then form an hypothesis about the behavior of the model on certain inputs. Along these lines, to detect a model’s reliance on spurious signals with feature attributions, one of the following must occur: i) The practitioner should inspect attributions for inputs that contain the spurious signal and notice that the spurious signal constitute the key feature for a model’s output decision or ii) The practitioner should inspect attributions for inputs that do not contain the spurious signal and notice `an issue’ with these attributions. In addition, the output decision to be explained has to also be the class for which the spurious signal encodes for. For example, in our experiments the hospital tag encodes for the pre-puberty class. These two settings are exactly what we measure with the SSIM metric in our analysis for feature attributions.

\paragraph{Concept Activation} these classes of approaches measure the dependence of a particular class to a user provided concept. For example, one could measure the pre-puberty class’ dependence on the hospital tag concept. As part of this approach, the hospital tag is user defined, and it is up to the practitioner to decide whether to test this concept. Other concepts include low frequency signals, patches of the image, color dimensions, or even any conceivable high-level concept the practitioner is interested in. For example, in the bone-age model, the concepts DIP, PIP, MCP, Wrist etc, are user defined clinical concepts that we chose and which are well specified for this setting.

For each concept of interest, the practitioner collects input examples that have this concept. For example, to test a model’s dependence on a hospital tag, we collect images that all have hospital tags in them. Given this collection of examples, the TCAV approach can then be used to calculate a score, TCAV score, that corresponds to the sensitivity of a particular output class to the input concept. The key insight that makes the TCAV approach susceptible to the limitations we point out are 1) it is up to the practitioner to decide which concept to test, and 2) our findings suggests that unless the spurious signal is explicitly tested, the TCAV scores for other concepts do not indicate that a model is reliant on a spurious signal. We note that even though we show all the concept scores in a single bar chart, these concepts are actually tested independently.

\paragraph{Training Point Ranking} These class of approaches ranks all training points by influence on the test loss of an input. Qualitatively, if a training point has a high influence on the test loss of an input, it means that if the model were re-trained without that training point, the test loss of that test point would change significantly. Intuitively, the most highly ranked training points for a given test input should also be points for which the model relies on semantically similar features. For example, one would expect the most highly ranked training points for a pre-puberty test-input to be other pre-puberty inputs as well (assuming a model has learned the right semantic features)

To use this approach to identify spurious signals, one would have to inspect an input that contains the spurious signal and further notice that the top ranked training inputs also include the training signal. Specifically, to test that a model is relying on the hospital tag to identify pre-puberty inputs one would have to use a test-input that has a hospital tag and then notice that all the top ranked inputs also have hospital tags. The crux of our argument is that the choice and decision to use a test-input with an hospital tag is the critical choice underlying whether the training point ranking methods can be used to effectively detect a model’s reliance on spurious signals. One measure of reliability for an influence function approach is the ICM metric that we compute in the paper.

Taken together, in the scenario where a practitioner is handed a model and asked to assess whether it relies on any spurious signal, detecting a model’s reliance spurious signals requires that the practitioner know which specific signals to test ahead of time.

\section{Model Training \& Dataset Processing}
\label{appendix:modelsdatasets}

\paragraph{Bone Age Dataset} We consider the high stakes task of predicting the bone age category from a radiograph to one of five classes based on age: Infancy/Toddler, Pre-Puberty, Early/MiD Puberty, Late Puberty, and Post Puberty. This task is one that is routinely performed by radiologists and as been previously studied with a variety of DNN. The dataset we use is derived from the Pediatric Bone Age Machine learning challenge conducted by the radiological society of North America in 2017~\cite{halabi2019rsna}. The dataset consists of $12282$ training, $1425$ validation, and $200$ test samples. We resize all images to (299 by 299) grayscale images for model training. We note here that the training, validation, and test set splits correspond to similar splits used for the competition, so we retain this split.

\begin{wrapfigure}{r}{0.4\textwidth}
  \begin{center}
\includegraphics[page=1, scale=0.25]{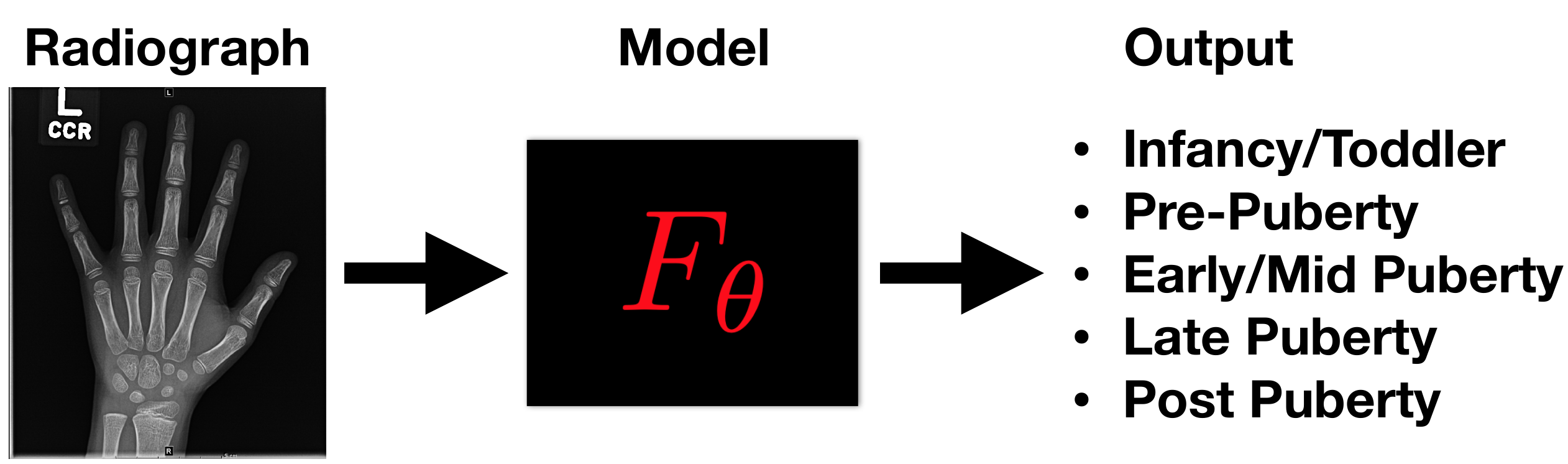}
\end{center}
\caption{Illustration of the Bone Age Task.}
\label{fig:boneagetagillustration}
\end{wrapfigure}

\paragraph{Knee Dataset}  We also consider the high stakes task of predicting the Kellgren and Lawrence (KL) grade of osteoarthritis based on Knee Xrays. The KL grade ranges from the integers $0$ to $4$; 5 classes, so we treat it as a classification task. The Knee Radiographs are obtained from the open source release by~\citet{chen2019fully} and consists of $5778$ training samples, $826$ validation samples, and $1656$ test samples. Again here, we follow~\citet{chen2019fully} data splits. The images were resized to be $299$ by $299$ pixels.

\paragraph{Dog Dataset} The third dataset that we use in this work is the dog breed classification dataset that is a combination of Stanford dogs dataset Khosla et al. (2011) and the
Oxford Cats and Dogs datasets Parkhi et al. (2012b) following~\cite{adebayo2020debugging}. Similarly,  we restrict our attention to 10 dog
classes: Beagle, Boxer, Chihuahua, Newfoundland, Saint Bernard, Pugs,
Pomeranian, Great Pyrenees, Yorkshire Terrier, Wheaten Terrier. Instead of the background spurious signal of\citet{adebayo2020debugging}, we focus instead on the three spurious signals tested in this work.

\paragraph{Models} We consider two different kinds of models: i) a small vanilla DNN based on~\cite{raghu2019transfusion}, and a Resnet-50 model. The small DNN consists of: conv-relu-batchnorm-maxpooling operation successively, and two fully connected layers at the end. All convolutional kernels have stride $1$, and kernel size $5$. We train this model with SGD with momentum (set to $0.9$) and an initial learning rate of $0.01$. We use a learning rate scheduler that decays the learning rate every $10$ epochs by $\gamma=0.1$.

\paragraph{Hyper-Parameter Tuning for Model Training} We used the Ray Tune library for hyper-parameter tuning of all the models used in this work. For the ResNet-50, we tuned with Ray, but the best performing models retained the default parameter settings. In the case of the Small DNN model, we tune the batch size, and learning rate with the validation set.

\paragraph{Compute} We perform all of our experiments on a VM with 60GB of RAM and a k80 GPU on Google Cloud.

\paragraph{A Note about random seeds and Model Runs.} We train $5$ models each (different random seeds) for each category and in the following tables the average performance metrics for these models. We found that the standard error of the mean for typically between $0.01-0.05$, so we omit these in the tables to improve readability.

\section{Additional Results: Feature Attributions}
\label{appendix:featureattributions}
In a series of subsequent figures with this appendix. We show additional saliency visualizations for settings considered in the main text. Here show a single visualization for the bone age setting since we found that the saliency attributions for such setting were quite faint. Overall we obtain similar results between the knee and bone age datasets, so we opt to show the bone age visualizations here instead.

\begin{figure*}[ht]
\centering
\includegraphics[page=8,scale=0.21]{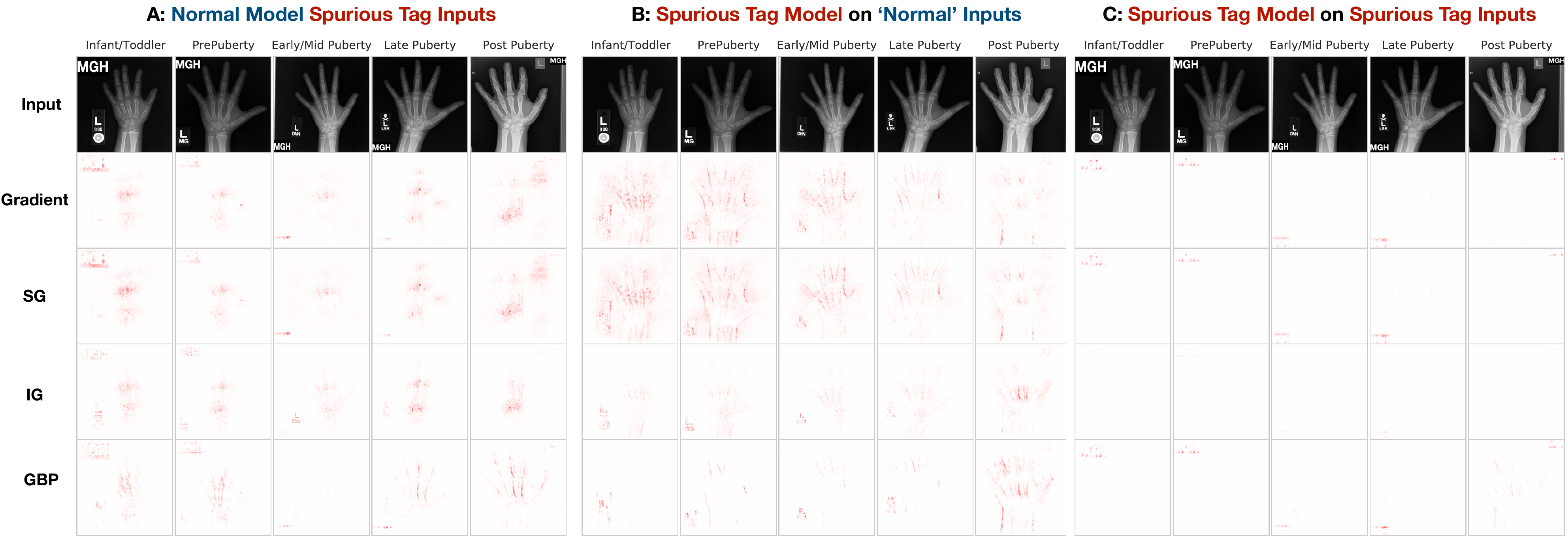}
\caption{\textbf{Detecting Spurious \textit{Tag}.} Here we show in A) Feature attributions for $5$ different inputs across the four feature attribution methods with a normal model but with spurious Tag inputs; B) Feature attributions on the same $5$ inputs as in (A), but \textbf{without} spurious Tag inputs with a model that has learned a spurious alignment between Pre-Puberty and Tag; C) Feature attributions on the same $5$ inputs as in (A), but \textbf{with} the spurious Tag inputs with a model that has learned a spurious alignment between Pre-Puberty and Tag.}
\label{fig:saliencyspurioustagmodel}
\end{figure*}

\begin{figure*}[ht]
\centering
\includegraphics[page=9,scale=0.21]{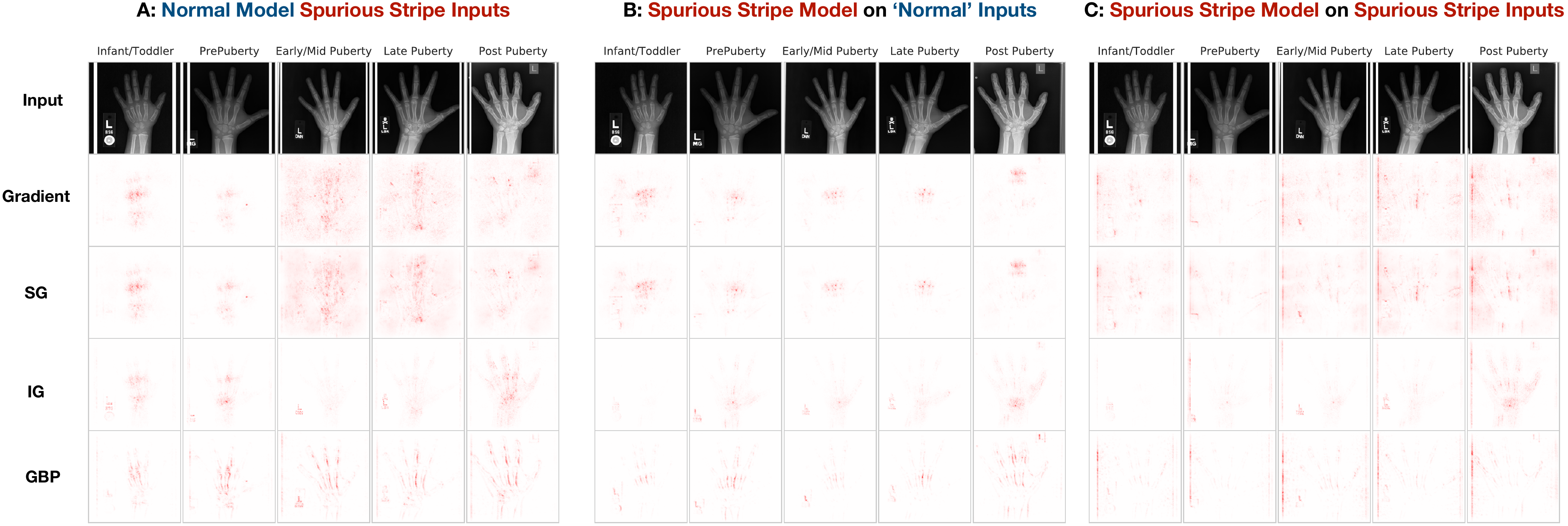}
\caption{\textbf{Detecting Spurious \textit{Stripe}.} Here we show in A) Feature attributions for $5$ different inputs across the four feature attribution methods with a normal model but with spurious Stripes inputs; B) Feature attributions on the same $5$ inputs as in (A), but \textbf{without} the spurious Stripe with a model that has learned a spurious alignment between Pre-Puberty and Stripe; C) Feature attributions on the same $5$ inputs as in (A), but \textbf{with} the spurious Stripe with a model that has learned a spurious alignment between Pre-Puberty and Stripe.}
\label{fig:saliencyspuriousstripemodel}
\end{figure*}

\begin{figure*}[ht]
\centering
\includegraphics[page=10,scale=0.21]{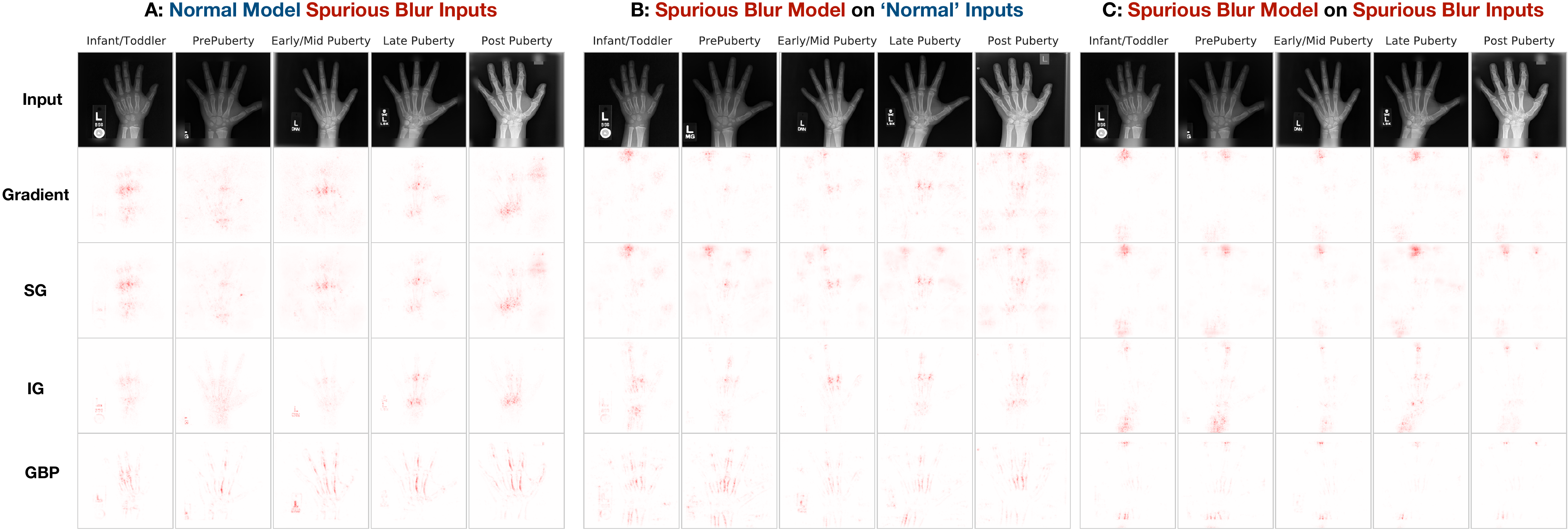}
\caption{\textbf{Detecting Spurious \textit{Blur}.} Here we show in A) Feature attributions for $5$ different inputs across the four feature attribution methods with a normal model but with spurious Blur inputs; B) Feature attributions on the same $5$ inputs as in (A), but \textbf{without} the spurious Blur with a model that has learned a spurious alignment between Pre-Puberty and Blur; C) Feature attributions on the same $5$ inputs as in (A), but \textbf{with} the spurious blur with a model that has learned a spurious alignment between Pre-Puberty and Blur.}
\label{fig:saliencyspuriousblurmodel}
\end{figure*}

\begin{figure*}[ht]
\centering
\includegraphics[scale=0.5]{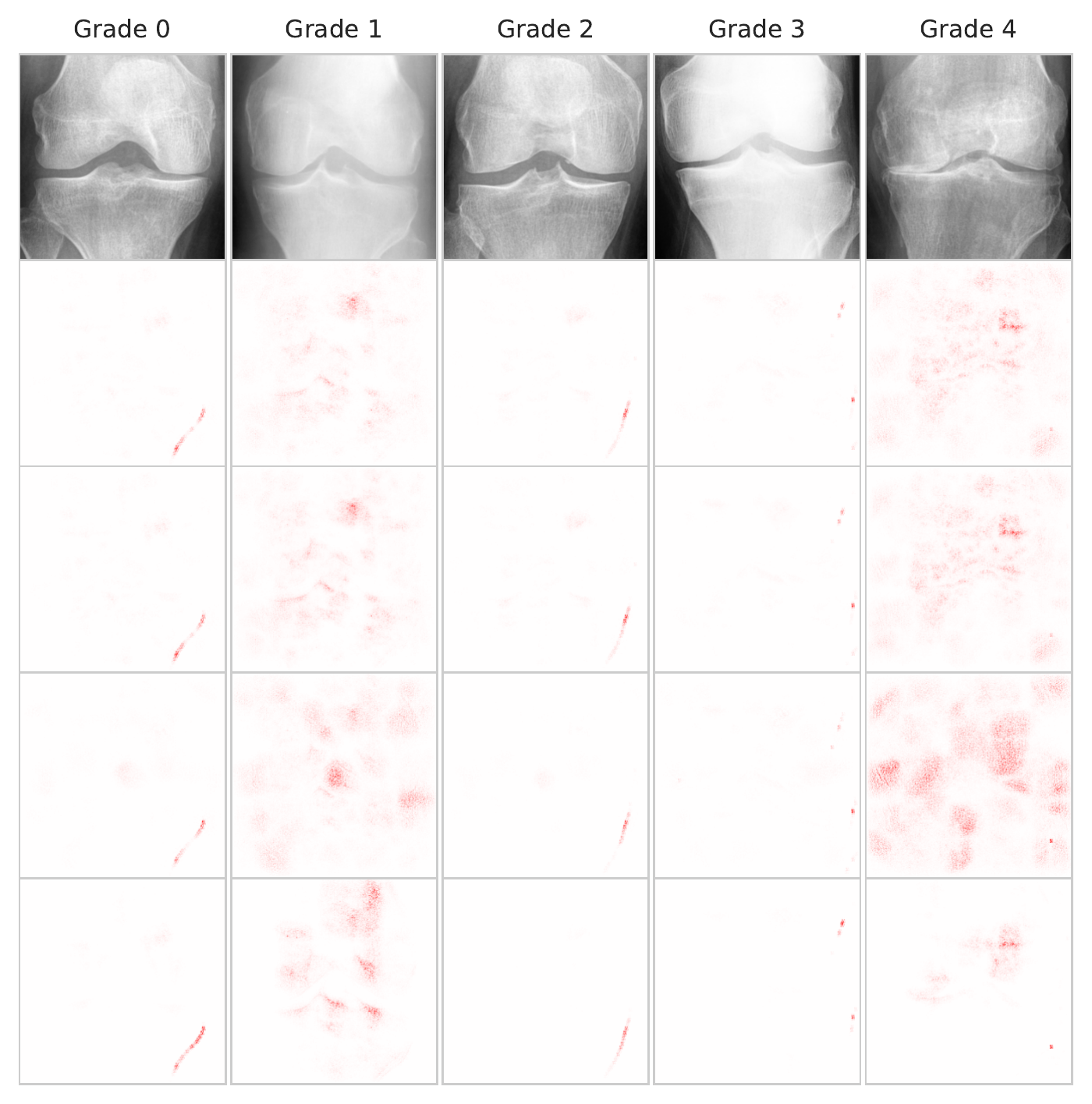}
\caption{Spurious Images for Spurious Knee Model Examples }
\label{fig:appendixnormalmodelspuriousblur}
\end{figure*}

\begin{figure*}[ht]
\centering
\includegraphics[scale=0.3]{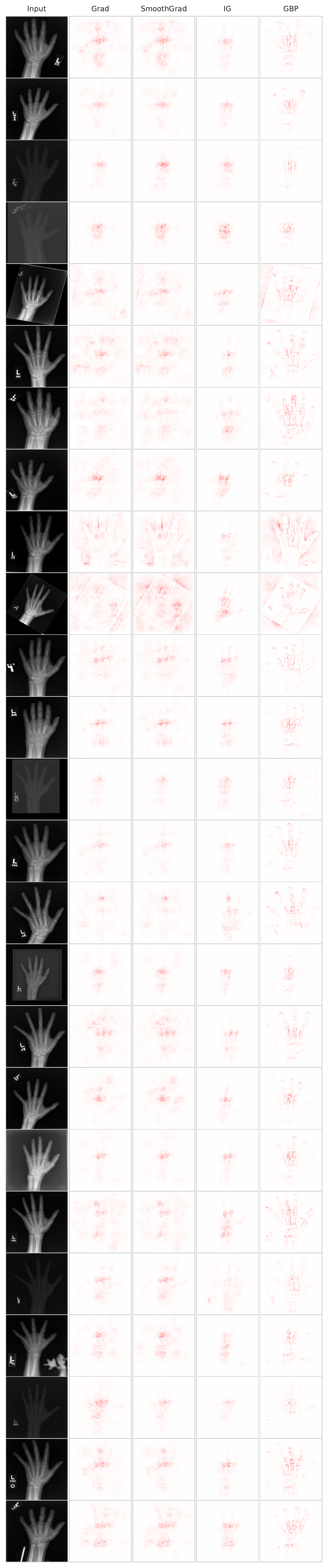}
\caption{Normal Model on Normal Images}
\label{fig:appendixnormalmodelspuriousblur}
\end{figure*}

\begin{figure*}[ht]
\centering
\includegraphics[scale=0.3]{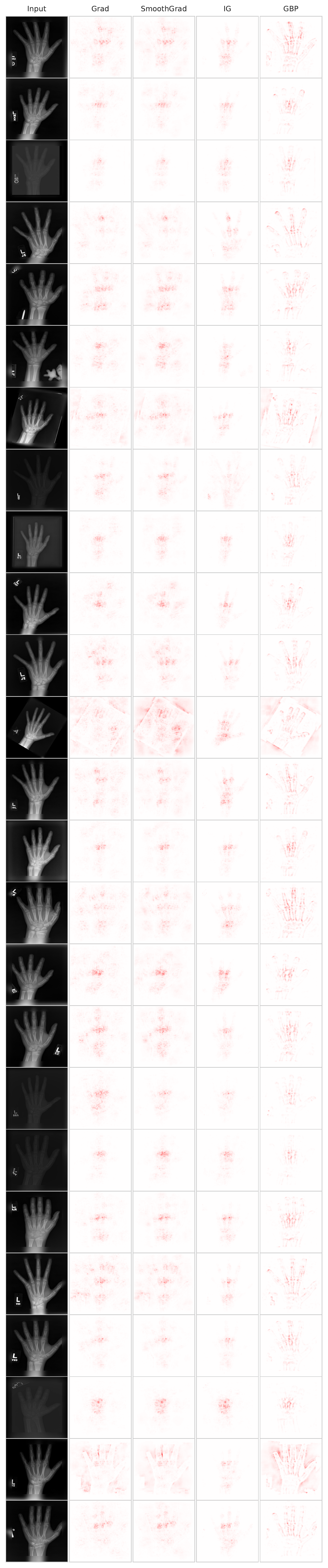}
\caption{Normal Model on Spurious Blur Images}
\label{fig:appendixnormalmodelspuriousblur}
\end{figure*}

\begin{figure*}[ht]
\centering
\includegraphics[scale=0.3]{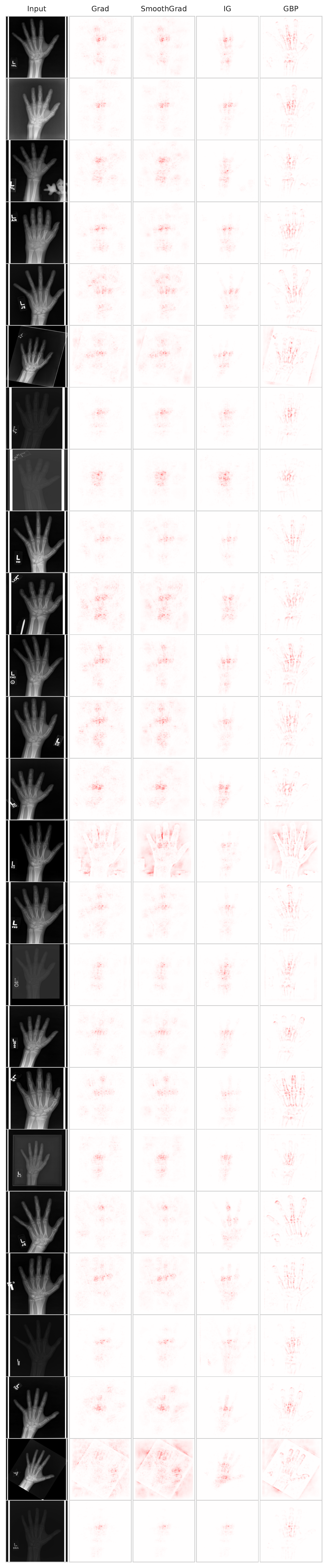}
\caption{Normal Model on Spurious Stripe Images}
\label{fig:appendixnormalmodelspuriousblur}
\end{figure*}

\begin{figure*}[ht]
\centering
\includegraphics[scale=0.3]{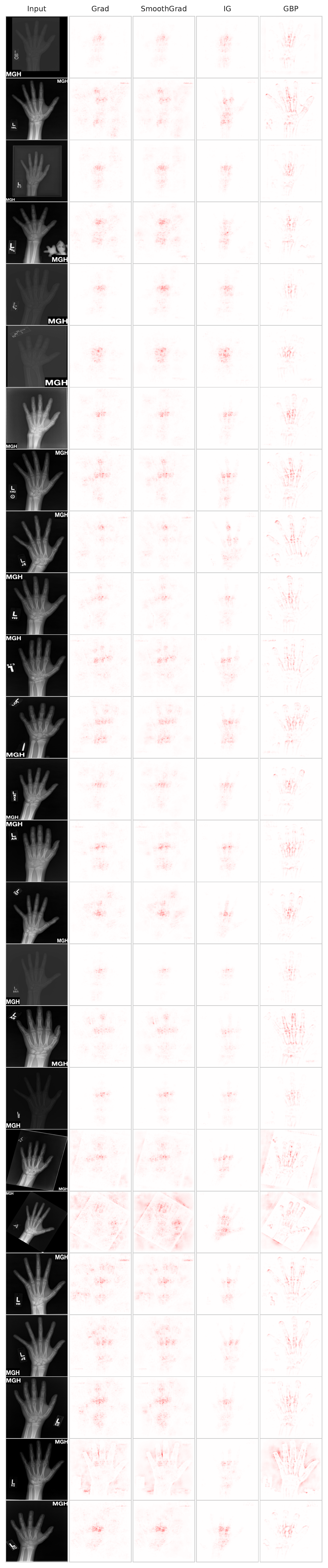}
\caption{Normal Model on Spurious Tag Images}
\label{fig:appendixnormalmodelspuriousblur}
\end{figure*}

\begin{figure*}[ht]
\centering
\includegraphics[scale=0.3]{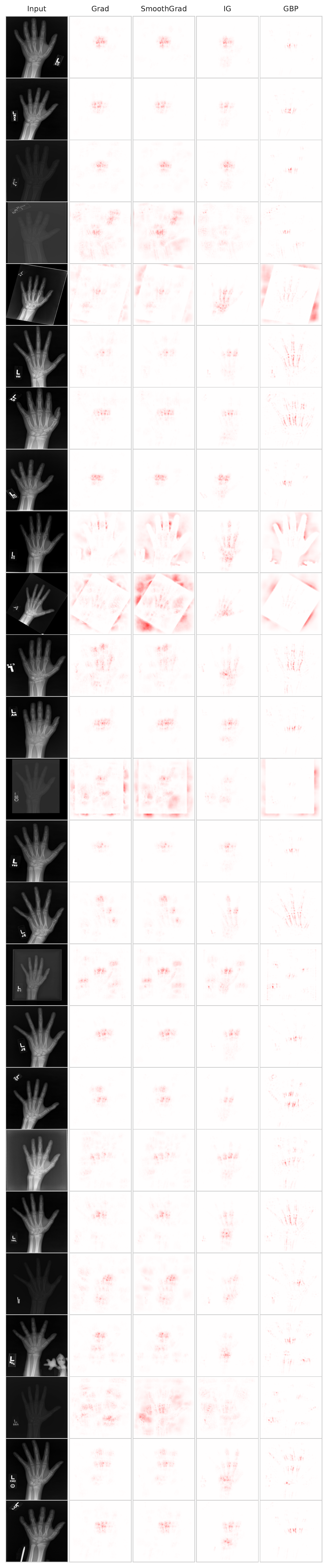}
\caption{Spurious Stripe Model on Normal Images}
\label{fig:appendixnormalmodelspuriousblur}
\end{figure*}

\begin{figure*}[ht]
\centering
\includegraphics[scale=0.3]{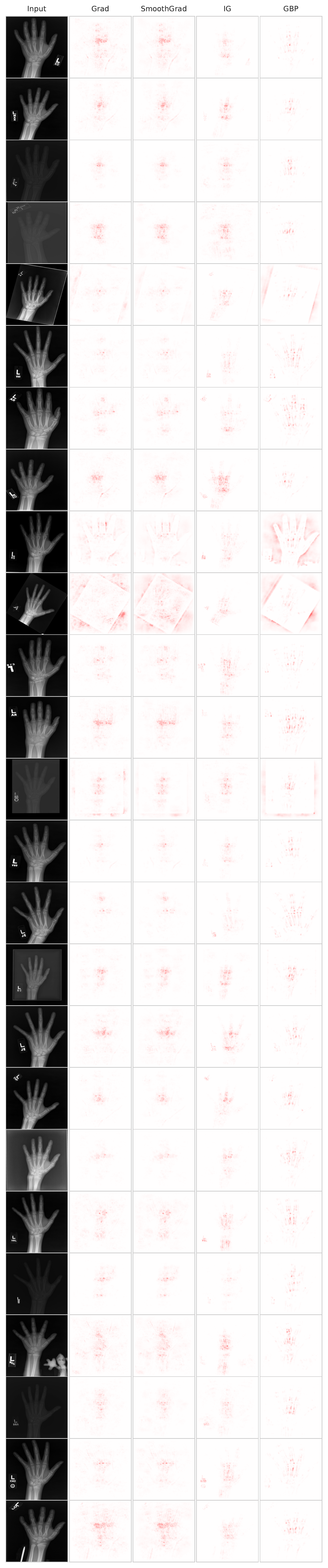}
\caption{Spurious Tag Model on Normal Images}
\label{fig:appendixnormalmodelspuriousblur}
\end{figure*}

\begin{figure*}[ht]
\centering
\includegraphics[scale=0.3]{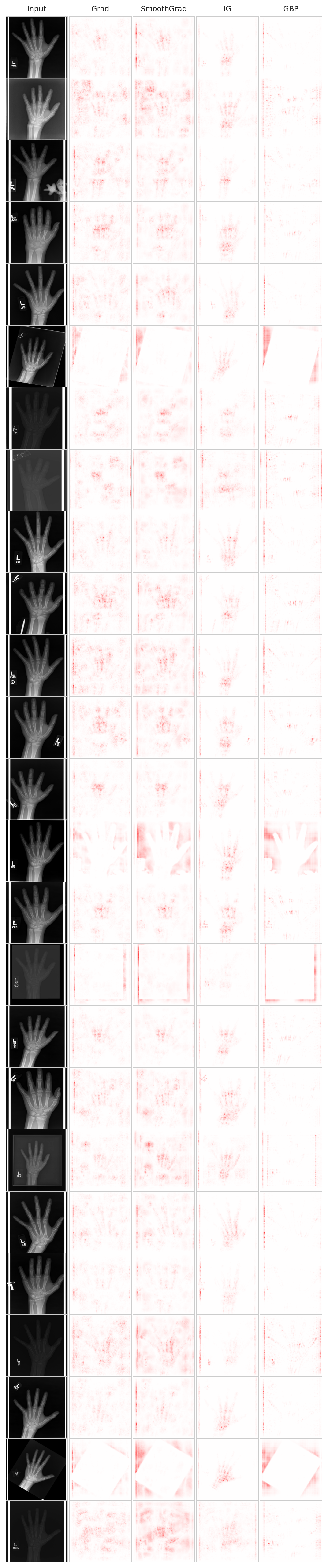}
\caption{Spurious Stripe Model on Spurious Stripe Images}
\label{fig:appendixnormalmodelspuriousblur}
\end{figure*}

\begin{figure*}[ht]
\centering
\includegraphics[scale=0.3]{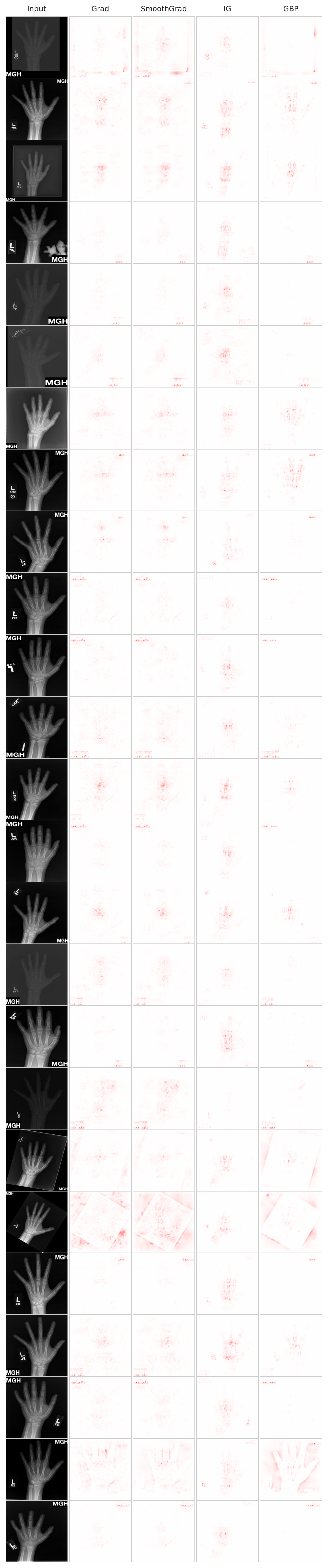}
\caption{Spurious Tag Model on Spurious Tag Images}
\label{fig:appendixnormalmodelspuriousblur}
\end{figure*}

\section{Additional Results: Concepts Approaches}
\label{appendix:concept}

\begin{figure*}[ht]
\centering
\includegraphics[scale=0.5]{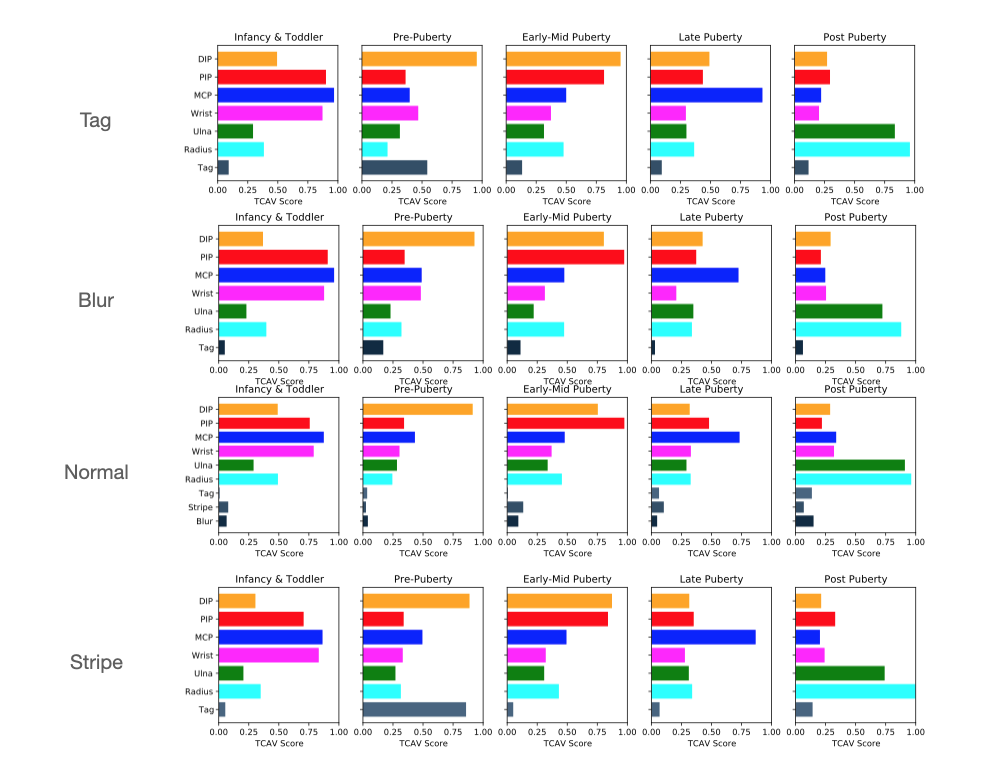}
\caption{Results for Concept Ranking for another run of a normal and spurious bone age models. }
\label{fig:appendixconceptnormalandspurious}
\end{figure*}

\paragraph{Normal Model compared to radiologist Concept Ranking} For each of the $100$ re-runs, we computed the rank correlation between the concept ranking for the normal model and rankings provided by a board certified radiologist.  

\begin{figure*}[ht]
\centering
\includegraphics[scale=0.5]{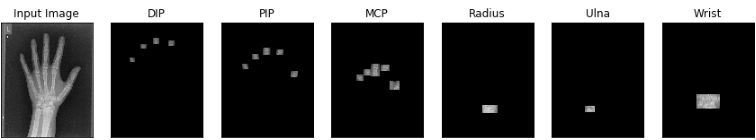}
\caption{\textbf{Concept Partition for Bone Age Example.} Here we show how we partition a single instance into its constituent clinical concept components.}
\label{fig:conceptpartition}
\end{figure*}

\paragraph{Concepts for Dogs and Bone Age.} We consider TCAV for the bone age and dog breeds task. For bone age, we choose as concepts the partitions of a hand, which correspond to the parts of the hand a radiologist would inspect to ascertain the age from a radiograph. We show, in Figure~\ref{fig:conceptpartition}, a representation of these concepts. In the dog breeds classification task, consider concepts related floppiness of the ear, erectness of the ear, the dog head type, and color.

We manually collect data on the dog breed concept. Here we showed Amazon mechanical Turkers an image from each class (10 classes) and then asked them to indicate whether each specific dog class had floppy ears or not, erect ears or not, single colored or not etc. We normalize each attribute to be binary and apply each concept broadly across each class.

\paragraph{Discussion on ACE}. The ACE concept approach segments an image and uses the image segments as concepts as part of a TCAV pipeline. In an experimental analysis, we segment the spurious images using the various segmentation algorithms in algorithm to identify how often the spurious signal is selected as a distinct segment. We found segmentation to be ineffective in the stripe and blur settings. For the spurious tag, we found segmentation to be effective in only 25 percent of the examples tested, which suggests that the underlying segmentation algorithm as part of ACE might be ineffective for detecting hidden spurious signals.

\section{Additional Results: Influence Functions}
\label{appendix:saliency}

\begin{figure*}[ht]
\centering
\includegraphics[page=12,scale=0.25]{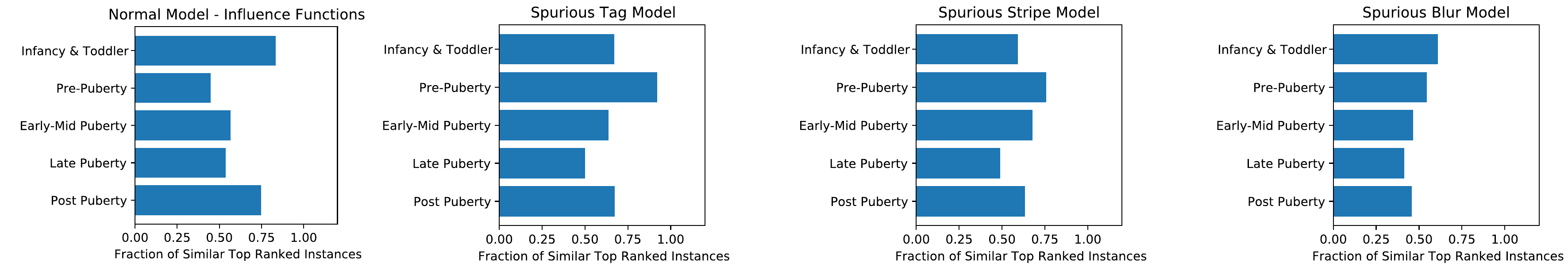}
\caption{\textbf{Influence Function Results.} Here we show the \textit{identical class metric (ICM)}~\citep{hanawa2020evaluation} for a normal model and models that rely on the three spurious signals tested in this work. ICM measures what fractions of the top training inputs for a given test example belong to the same class as the true class of the test example in question.}
\label{fig:influencefunctionsspurious}
\end{figure*}

\begin{figure*}[!h]
\centering
\includegraphics[scale=0.5]{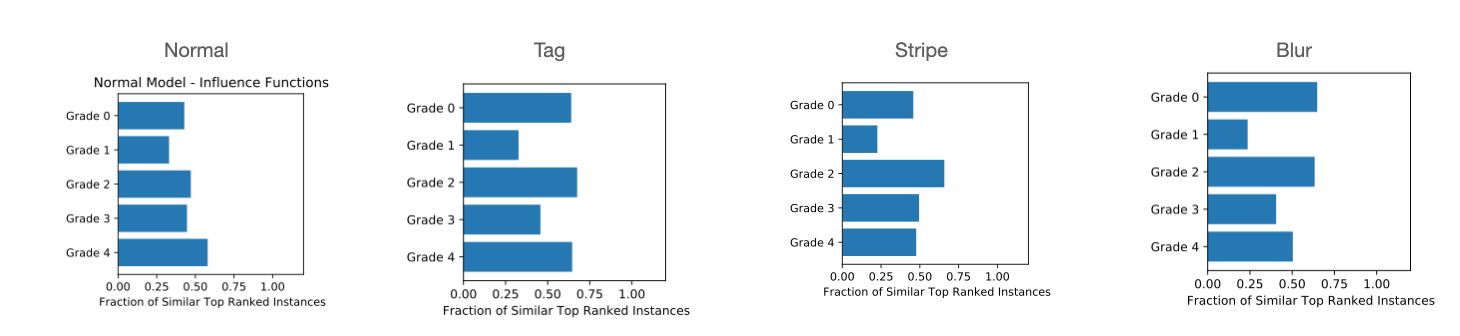}
\caption{Influence Functions Results for Knee Settings. }
\label{fig:appendixinfluencekneesettings}
\end{figure*}

In Figure~\ref{fig:appendixinfluencekneesettings}, we show the influence metric that we computed in the main text for the small DNN model for the Knee Arthritis prediction. In this task, the spurious signal is always aligned with the Grade 2 class. Here again, we see that in the spurious settings each metric actually improves.

\section{User Study Discussion}
\label{appendix:userstudy}

In this section we present additional results of the user study along with break down of some demographic information about the participants.

\begin{table}[!h]
\centering
\rowcolors{1}{}{lightgray}
\begin{tabular}{ccccc}
 \toprule
 Gender & Counts\\
 \midrule
Male & 130\\ 
Female & 62\\
N/A & 8\\ 
 \bottomrule
\end{tabular}
\caption{\textbf{Gender Breakdown across participants}.}
\label{tab:userstudy}
\end{table}

\begin{table}[!h]
\centering
\rowcolors{1}{}{lightgray}
\begin{tabular}{ccccc}
 \toprule
ML Background & Counts\\
 \midrule
ML Researcher & 30\\ 
ML Practitioner & 21\\
Major Familiarity with ML & 50\\ 
Limited Familiarity & 47\\
No Familiarity & 52\\ 
 \bottomrule
\end{tabular}
\caption{\textbf{Machine Learning Experience Breakdown across participants}.}
\label{tab:userstudy}
\end{table}

\begin{figure*}[ht]
\centering
\includegraphics[scale=0.50]{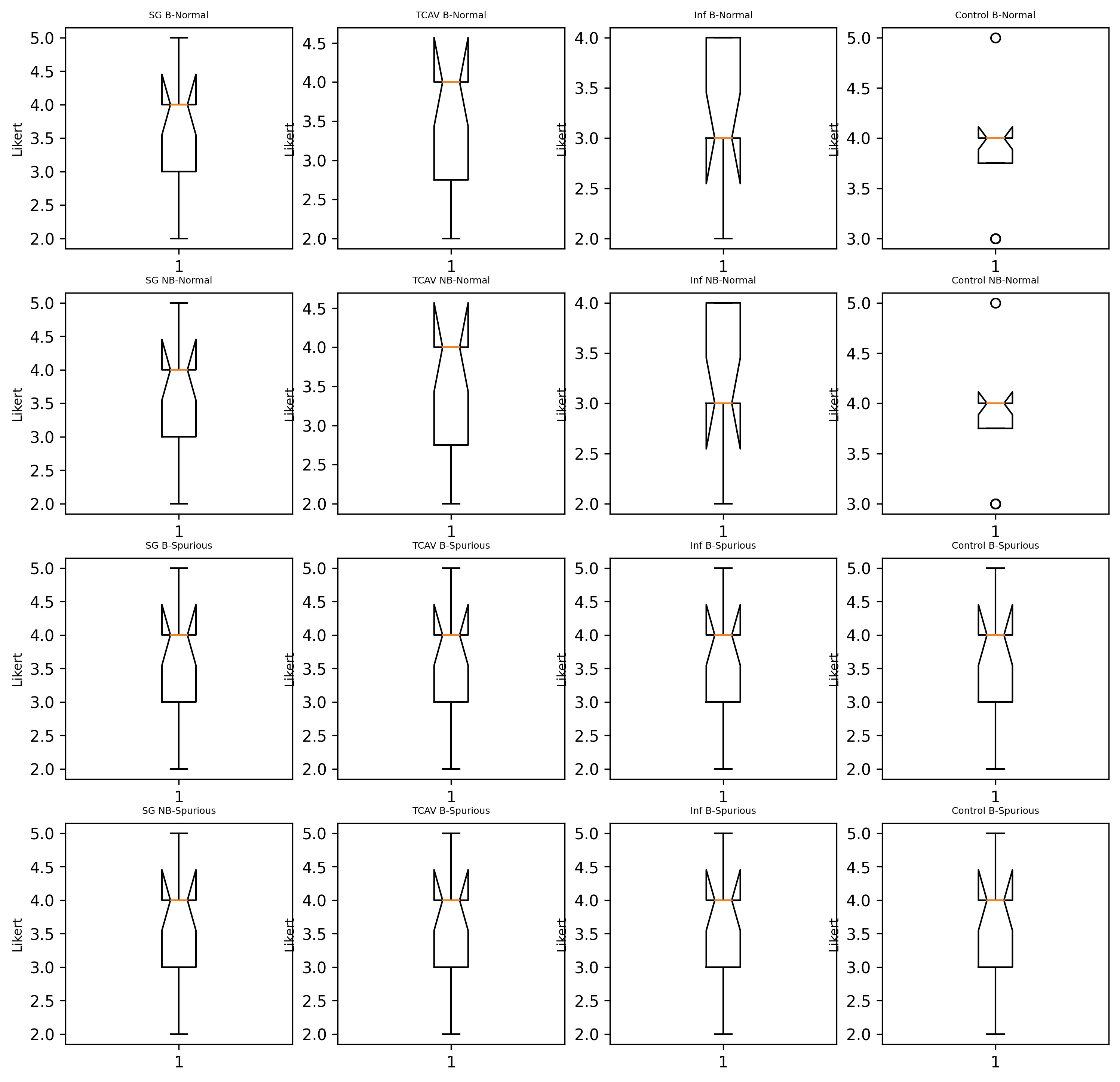}
\caption{\textbf{Box plots of User responses.} }
\label{fig:saliencymastercompilation}
\end{figure*}

\newpage

We now provide additional tables for the the feature attribution metrics from the main text.

\begin{table*}[th]
\centering
\rowcolors{1}{}{lightgray}
\resizebox{\linewidth}{!}{\begin{tabular}{cccccccccccccccc}
\hline
\textbf{Metric} & \textbf{Gradient} & \textbf{SmoothGrad} & \textbf{Integrated Gradients} & \textbf{Guided BackProp} \\
\hline
SISM-GT & 0.68 & 0.74 & 0.62 & 0.92 \\
SEM & 0.0091 & 0.017 & 0.019 & 0.0038\\
NISM-SISM & 0.0013 & 8.8e-4 & 0.07 & 0.37\\
SEM & 0.0012 & 0.0082 & 0.076 & 0.059\\
\hline
\end{tabular}}
\caption{\textbf{Metrics for Bone Age Spurious Tag Model.}}
\label{tab:boneagespurioustag}
\end{table*}

\begin{table*}[th]
\centering
\rowcolors{1}{}{lightgray}
\resizebox{\linewidth}{!}{\begin{tabular}{cccccccccccccccc}
\hline
\textbf{Metric} & \textbf{Gradient} & \textbf{SmoothGrad} & \textbf{Integrated Gradients} & \textbf{Guided BackProp} \\
\hline
SISM-GT & 0.78 & 0.68 & 0.59 & 0.89 \\
SEM & 0.0011 & 0.012 & 0.019 & 0.0088\\
NISM-SISM & 0.0013 & 7.2e-5 & 0.17 & 0.46\\
SEM & 0.0016 & 0.0032 & 0.026 & 0.029\\
\hline
\end{tabular}}
\caption{\textbf{Metrics for Bone Age Spurious Stripe Model.}}
\label{tab:boneagespurioustag}
\end{table*}

\begin{table*}[th]
\centering
\rowcolors{1}{}{lightgray}
\resizebox{\linewidth}{!}{\begin{tabular}{cccccccccccccccc}
\hline
\textbf{Metric} & \textbf{Gradient} & \textbf{SmoothGrad} & \textbf{Integrated Gradients} & \textbf{Guided BackProp} \\
\hline
SISM-GT & 0.7 & 0.84 & 0.49 & 0.92 \\
SEM & 0.001 & 0.02 & 0.06 & 0.008\\
NISM-SISM & 0.0013 & 2.4e-4 & 0.21 & 0.86\\
SEM & 0.006 & 0.002 & 0.06 & 0.09\\
\hline
\end{tabular}}
\caption{\textbf{Metrics for Bone Age Spurious Blur Model.}}
\label{tab:boneagespurioustag}
\end{table*}

\begin{table}[!h]
\centering
\caption{Performance metrics for each attribution method across tasks for the Tag Setting.}
\begin{tabular}{ccccccccccccc}
\toprule
Method &  \multicolumn{4}{c}{Bone Age} & \multicolumn{4}{c}{Knee} & \multicolumn{4}{c}{Dog Breed}\\
\midrule
{}   & Grad   & SG    & IG   &  GBP & Grad & SG & IG & GBP & Grad & SG & IG & GBP \\
K-SSD   &  0.65 & 0.66   & 0.67  & 0.81 & 0.51 & 0.49 & 0.47 & 0.76 & 0.71 & 0.76   & 0.79  & 0.88\\
CCM  &  0.37 & 0.39   & 0.35 & 0.75 & 0.32 & 0.33 & 0.35 & 0.66 & 0.42 & 0.41   & 0.39 & 0.64\\
FAM   &  0.51  & 0.55   & 0.53  & 0.68  & 0.46 & 0.47 & 0.45 & 0.69 & 0.59  & 0.64 & 0.68  & 0.73\\
\bottomrule
\end{tabular}
\label{tab:measurestag}
\end{table}

\begin{table}[!h]
\centering
\caption{Performance metrics for each attribution method across tasks for the Blur Setting.}
\begin{tabular}{ccccccccccccc}
\toprule
Method &  \multicolumn{4}{c}{Bone Age} & \multicolumn{4}{c}{Knee} & \multicolumn{4}{c}{Dog Breed}\\
\midrule
{}   & Grad   & SG    & IG   &  GBP & Grad & SG & IG & GBP & Grad & SG & IG & GBP \\
K-SSD   &  0.21 & 0.20   & 0.19  & 0.13 & 0.13 & 0.18 & 0.17 & 0.31 & 0.29 & 0.30 & 0.31 &0.35\\
CCM  &  0.28 & 0.29   & 0.24 & 0.64 & 0.23 & 0.22 & 0.27 & 0.67 & 0.38 & 0.33 & 0.35 & 0.71\\
FAM   &  0.48  & 0.49 & 0.47  & 0.51  & 0.36 & 0.38 & 0.33 & 0.58 & 0.55  & 0.56 & 0.47  & 0.73\\
\bottomrule
\end{tabular}
\label{tab:measuresblur}
\end{table}